\documentclass{article} 
\usepackage{main,times}

\usepackage{amsmath,amsfonts,bm}

\def\eqref#1{equation~\ref{#1}}

\def\1{\bm{1}}

\DeclareMathAlphabet{\mathsfit}{\encodingdefault}{\sfdefault}{m}{sl}
\SetMathAlphabet{\mathsfit}{bold}{\encodingdefault}{\sfdefault}{bx}{n}

\usepackage{hyperref}
\usepackage{url}
\usepackage{wrapfig}
\usepackage[most]{tcolorbox}
\usepackage{booktabs}
\usepackage{amsmath}
\usepackage{graphicx} 
\usepackage{booktabs}
\usepackage{multirow} 
\usepackage{enumitem}
\usepackage{subcaption} 
\usepackage{threeparttable}
\usepackage{tabularx}
\usepackage{titletoc}
\usepackage{float}

\title{Automated Optimization Modeling via a Localizable Error-Driven Perspective}

\author{\textbf{Weiting Liu}$^{1}$ \quad
\textbf{Han Wu}$^{2}$ \quad
\textbf{Yufei Kuang}$^{3}$ \quad
\textbf{Xiongwei Han}$^{2}$ \quad 
\textbf{Tao Zhong}$^{2}$ \\[2pt]
\textbf{Jianfeng Feng}$^{1}$ \quad
\textbf{Wenlian Lu}$^{1}$\thanks{Correspondence: wenlian@fudan.edu.cn} \\[4pt]
$^{1}$Fudan University \quad
$^{2}$Huawei Noah’s Ark Lab \quad
$^{3}$University of Science and Technology of China
}

\iclrfinalcopy 
\begin{document}

\maketitle

\begin{abstract}
Automated optimization modeling via Large Language Models (LLMs) has emerged as a promising approach to assist complex human decision-making. While post-training has become a pivotal technique to enhance LLMs' capabilities in this domain, its effectiveness is severely constrained by the scarcity and underutilization of high-quality training data. However, through a detailed profiling of error patterns across various problem-response pairs drawn from post-training, we identify two fundamental limitations of existing automated optimization modeling approaches: (L1) the \textit{sparsity} of error-specific problems and (L2) the \textit{sparse rewards} associated with difficult problems. We demonstrate that these limitations can result in suboptimal performance in domain-specific post-training for LLMs. To tackle the above two limitations, we propose a novel error-driven learning framework---namely, auto\textbf{m}ated opt\textbf{i}mization modeli\textbf{n}g via a localizable error-\textbf{d}riven perspective (MIND)---that customizes the whole model training framework from data synthesis to post-training. MIND is based on our key observation of the unique \textbf{\textit{localizable}} patterns in error propagation of optimization modelings, that is, modeling errors may remain localized to specific semantic segments and do not propagate throughout the entire solution. Thus, in contrast to holistic reasoning tasks such as mathematical proofs, MIND leverages the construction of a focused, high-density training corpus and proposes \textbf{D}ynamic Supervised \textbf{F}ine-Tuning \textbf{P}olicy \textbf{O}ptimization (DFPO) to tackle difficult problems through localized refinement. Its appealing features include that (1) it generates targeted, error-aware training problems that achieve superior sample efficiency, and (2) it ensures a coherent and structured learning progression for stable and effective reinforcement learning on difficult problems. Experiments on six benchmarks demonstrate that MIND \textit{consistently} outperforms all the state-of-the-art automated optimization modeling approaches. Furthermore, we open-source a new training dataset, MIND-Train, and a new benchmark, MIND-Bench, for the automated optimization modeling research community.
\end{abstract}

\section{Introduction}
\label{sec:intro}
Advances in computational power and algorithmic techniques have made optimization a fundamental tool across engineering~\citep{antoniou2007practical}, economics~\citep{intriligator2002mathematical}, logistics~\citep{bartolacci2012optimization}, manufacturing~\citep{rao2010advanced}, and artificial intelligence~\citep{kingma2014adam}, enabling more intelligent and data-driven decision-making. Optimization seeks values for decision variables that maximize or minimize an objective function while satisfying a set of constraints. Optimization modeling formalizes complex real-world problems into mathematical representations by defining variables, objectives, and constraints, allowing state-of-the-art solvers such as Gurobi~\citep{gurobi}, PySCIPOpt~\citep{pyscipopt}, and CPLEX~\citep{cplex} to efficiently compute solutions. Recently, the emergence of Large Language Models (LLMs) has opened a new avenue for automated optimization modeling, enabling the translation of natural language problem descriptions directly into mathematical formulations and executable solver code. Although automated optimization modeling cannot guarantee complete accuracy, their ability to rapidly generate candidate formulations to support human experts in optimization modeling is nonetheless of substantial practical value.

    \begin{figure}
        \centering
        \includegraphics[width=1.0\linewidth]{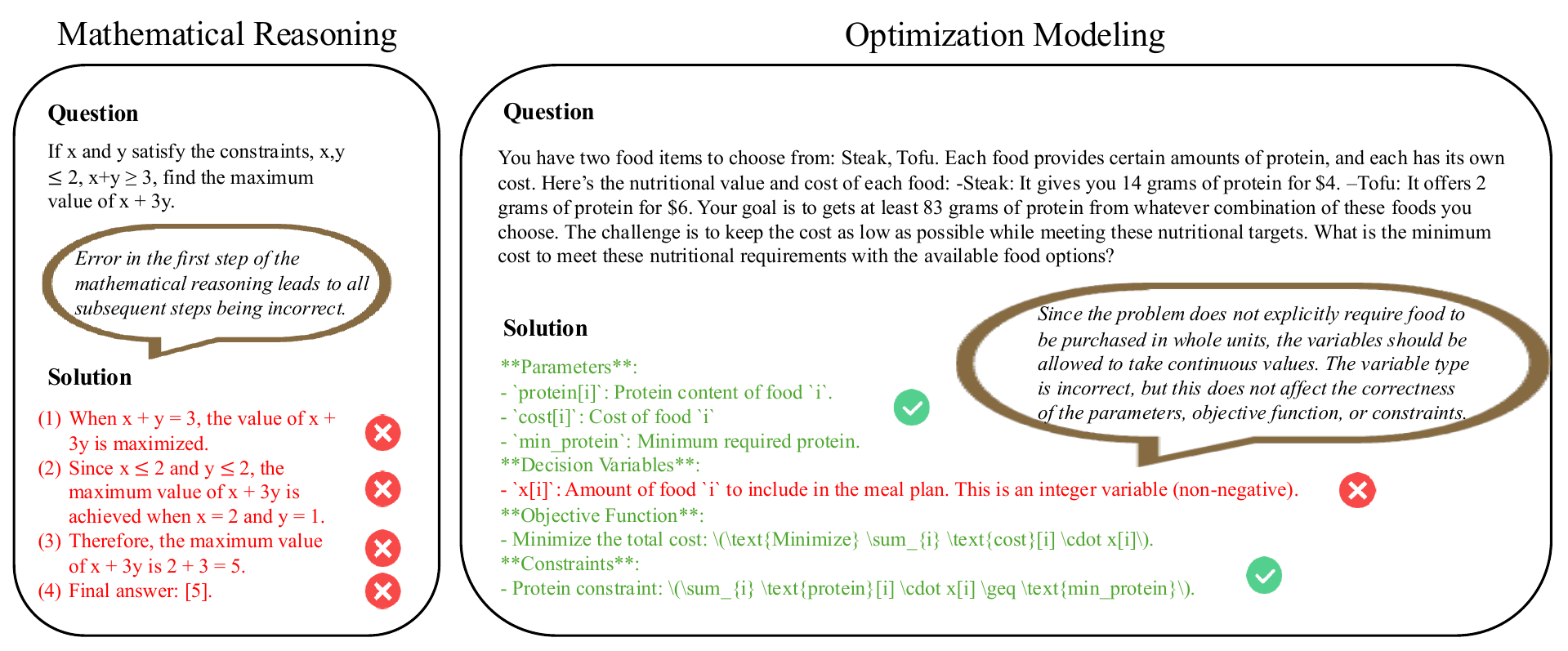}
        \caption{Illustration of the difference between mathematical reasoning and optimization modeling.}
        \label{fig:optimization_model_math_diff}
    \end{figure}

Recently, many general post-training techniques have been successfully adapted to improve the performance of automated optimization modeling.
A range of studies, such as ORLM \citep{huang2025orlm}, ReSocratic~\citep{yang2024optibench}, Step-Opt~\citep{wu2025step} and OptMATH~\citep{lu2025optmath}, adopt the paradigm of first synthesizing new data and subsequently fine-tuning models on the generated data. Another line of research, including LLMOPT~\citep{ethayarajh2024model} and SIRL~\citep{chen2023alpagasus}, explores the adaptation of reinforcement learning methods to this domain. For instance, SIRL introduces partial KL regularization and leverages solver feedback as a reward signal to update the model. A distinct line of methods focuses on test-time scaling (TTS), which effectively enhances model performance at inference without modifying the underlying parameters. Within this line, Chain-of-Experts~\citep{xiao2023chain} and OptiMUS~\citep{ahmaditeshnizi2023optimus} explore multi-agent systems, whereas Autoformulator~\citep{astorga2024autoformulation} leverages Monte-Carlo Tree Search.
However, progress in this field remains constrained by two major challenges: (1) High cost of generating high-quality data. Existing methods rely heavily on seed data and demonstrate limited generalization beyond the scope of that data. (2) Sparse reward signals. Representative approaches, such as SIRL, primarily use the correctness of the final outcome as the reward signal, which tends to be sparse, particularly for difficult problems.
However, our insight reveals that LLMs typically make errors only within a limited subset of optimization modeling formulations—such as those involving variables, constraints, or objectives—rather than across all components (as illustrated in Fig~\ref{fig:optimization_model_math_diff}). This observation suggests that the formulation of these factors exhibits relative independence, which in turn motivates us to exploit this characteristic in both the data generation and training stages.

In this work, we propose a novel error-driven learning framework---namely, auto\textbf{m}ated opt\textbf{i}mization modeli\textbf{n}g via a localizable error-\textbf{d}riven perspective (MIND) to address the aforementioned challenges. Specifically, MIND is a two-stage framework: (1) Motivated by our key observation of the unique localizable patterns in error propagation of optimization modeling, we propose an error-driven reverse data synthesis pipeline to construct a focused, high-density training corpus, MIND-Train, which captures common error patterns to support the post-training pipeline; (2) To mitigate the sparse reward problem arising from the limited capacity of the base model on difficult problems, we introduce a novel \textbf{D}ynamic Supervised \textbf{F}ine-tuning \textbf{P}olicy \textbf{O}ptimization method (DFPO) that dynamically corrects wrong responses while generating corrected responses that remain close to the distribution of the base model’s responses during the training stage. By leveraging this slight distributional discrepancy, we integrate the supervised fine-tuning (SFT) and reinforcement learning (RL) in a novel, stable, and effective manner for automated optimization modeling.

Our contributions are summarized as follows: (1) \textit{Conceptually}, through extensive empirical analysis, we observe a low error ratio in automated optimization modeling, highlighting a key difference from general mathematical problems. (2) \textit{Methodologically}, we propose a novel error-driven learning framework to customize the entire model training framework from data synthesis to post-training to address two challenges in automated optimization modeling: the sparsity of error-specific problems and the scarcity of learning signals on difficult problems. (3) \textit{Experimentally}, we evaluate MIND on six benchmarks, demonstrating that it outperforms state-of-the-art automated optimization modeling methods. (4) \textit{From a data perspective}, we open-source a new training dataset, MIND-Train, and a new benchmark, MIND-Bench, for the automated optimization research community.

\section{Related Work}
\label{sec:relatedwork}
\paragraph{Domain-specific Data Synthesis and Augmentation}

    Recently, data generation methods have followed two main directions: data augmentation, which enhances existing samples through transformations (including data labeling~\citep{khan2023q}, data reformation~\citep{dunlap2023diversify}, and co-annotation~\citep{li2023coannotating}), and data synthesis, which creates entirely new samples either from scratch or using generative models. With the advancements of LLMs~\citep{brown2020language}, data synthesis has made significant strides in both the quality and efficiency of synthetic data generation. General model distillation~\citep{chen2023alpagasus, eldan2023tinystories, li2023textbooks}, domain model distillation~\citep{lewkowycz2022solving, luo2023wizardmath}, and model self-improvement~\citep{maini2024rephrasing, wang2022self, zelikman2022star} have emerged as mainstream data synthesis methods. Benefiting from verifiable outputs, data synthesis methods in mathematics, such as those in~\citep{zelikman2022star, luo2023wizardmath}, generate diverse questions, answers, and more rationale corpora, which are preserved after verification. Similar to general mathematics, optimization modeling can also be verified using an optimizer solver. There are three common data synthesis and augmentation methods in this domain. ORLM~\citep{huang2025orlm} applies data augmentation to transform existing automated modeling instances and utilizes forward data synthesis to rephrase questions, subsequently employing LLMs to generate corresponding mathematical formulations. Step-Opt~\citep{wu2025step} employs iterative problem generation, evolving both complexity and scope, to systematically and effectively augment existing datasets. In contrast, Resocratic~\citep{yang2024optibench} proposes a reverse data synthesis approach that rephrases formulations and then leverages LLMs to generate the corresponding questions. Combining these methods, OptMATH~\citep{lu2025optmath} introduces bidirectional data synthesis, which first rephrases mathematical formulations, then uses LLMs to generate questions, and finally applies LLMs again to produce mathematical formulations. The two sets of mathematical formulations are then compared to ensure data quality. Although these data synthesis and augmentation methods have successfully applied general data synthesis and augmentation techniques to the automated modeling domain, they overlook the unique characteristics of automated optimization modeling data. This gap motivates the development of MIND.

\paragraph{Domain-specific Post-Training}

    The predominant post-training techniques can be broadly categorized into fine-tuning~\citep{ouyang2022training, lester2021power,luong2024reft}, alignment~\citep{kaufmann2024survey,bai2022constitutional,rafailov2023direct}, and reasoning~\citep{gou2023critic,jaech2024openai,guo2025deepseek}. By leveraging the verifiable answer characteristics in mathematics~\citep{hu2025open} and code generation~\citep{luo2025deepcoder}, Reinforcement Learning with Verifiable Rewards (RLVR) has made significant progress in addressing these complex reasoning problems. The success of the representative RLVR method Group Relative Policy Optimization (GRPO)~\citep{shao2024deepseekmath} has inspired increasing research on improving RLVR methods through techniques such as normalization, clipping, data filtering, and loss aggregation. Compared to Proximal Policy Optimization (PPO)~\citep{schulman2017proximal}, GRPO~\citep{shao2024deepseekmath} computes response-level advantages for prompts within a group, replacing the value function used in PPO to improve training efficiency. Based on GRPO, Decoupled Clip and Dynamic Sampling Policy Optimization (DAPO)~\citep{yu2025dapo} introduce four curated tricks: it decouples the upper and lower clipping ranges to encourage exploration and prevent entropy collapse, dynamically filters out samples where all responses are correct or incorrect to improve training efficiency and stability, aggregates losses at the token level to better handle long responses, and applies special reward shaping to control overlong or truncated responses. To address the training instability and inefficiency of the RLVR method, Guided Hybrid Policy Optimization (GHPO)~\citep{liu2025ghpo} explores the use of hints extracted from the ground-truth solution during the reinforcement learning process. Unlike these approaches, Value-model-based Augmented Proximal Policy Optimization (VAPO)~\citep{yue2025vapo} uses a value-model-based RLVR method and adds a negative log-likelihood loss for correctly sampled outcomes. Within the vertical domain of automated optimization modeling, ORLM~\citep{huang2025orlm}, Step-Opt~\citep{wu2025step}, Resocratic~\citep{yang2024optibench}, and OptMATH~\citep{lu2025optmath} investigate supervised fine-tuning (SFT), LLMOPT~\citep{jiang2024llmopt} explores Kahneman–Tversky Optimization, and SIRL~\citep{chen2025solver} examines RLVR. Although these approaches apply general post-training techniques to automated optimization modeling, they overlook its unique characteristics. Building on the progress of these methods, we emphasize that RLVR can effectively bridge the gap between general-purpose LLMs and the specific requirements of automated optimization modeling from an error-driven perspective.

\section{Preliminaries}
\label{sec:model}
\subsection{Automated Optimization Modeling}

    In general, optimization modeling entails a complex chain-of-thought~\citep{wei2022chain}, including problem analysis, extraction of key information to build a rationale, formulation of a mathematical model with variables, objective functions, and constraints, followed by translation into executable code. An automated optimization modeling instance is defined as a tuple $(q, o, a)$, where $q$ denotes the natural language description of the question, $o$ represents the corresponding reasoning path consisting of the rationale $\mathcal{Z}$, mathematical formulation $\mathcal{MF}$, and executable code $\mathcal{C}$, and $a$ is the resulting objective value. Thus, the corresponding training instance is expressed as $(q, a^*)$, where $a^*$ denotes the ground-truth objective of $q$. The problem of automated optimization modeling is to transform $q$ into $o$, such that an optimization solver can execute the code $\mathcal{C}$ contained in $o$ to compute the objective value $a$. The goal is to find a reasoning path $o$ that yields an objective value $a$ matching the ground-truth objective $a^*$, thereby corresponding to the correct optimization modeling. We formulate the automated optimization modeling problem as follows:

    \begin{align}
        \max_\theta \ & \mathbb{E}_{(q, a^*) \sim \mathcal{D}, o \sim \pi_\theta(\cdot | q), a \sim \mathrm{BS}(o)} 
        \left[ R(a, a^*) \right],
    \end{align}

    where $\mathcal{D}$, $\theta$ and $\mathrm{BS}$ denote the training dataset, the parameters of the target policy $\pi_{\theta}$ and the backbone solver, respectively. Given a question $q$, the policy $\pi_{\theta}$ produces a reasoning path $o$. The backbone solver, such as PySCIPOpt, takes the reasoning path $o$ as input, extracts the corresponding executable code $\mathcal{C}$, and outputs the objective value $a$. Finally, $a$ is compared with the ground truth $a^*$ to compute the reward $R$.

\subsection{Preliminary Results} \label{sec:motivation}

\begin{wrapfigure}{r}{0.49\textwidth}
    \vspace{-1mm}
    \centering
    \includegraphics[width=0.48\textwidth]{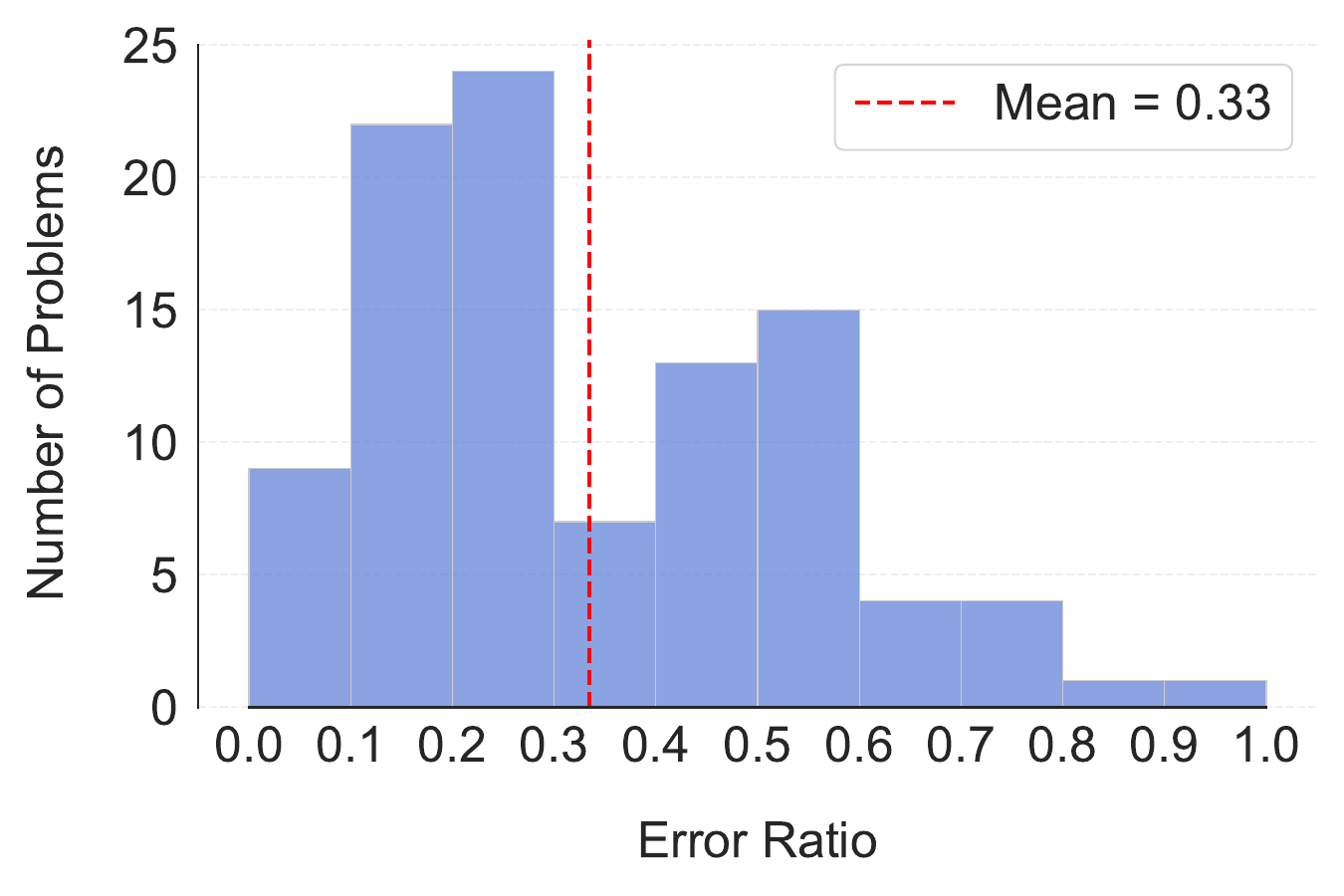}
    \vspace{-2.5mm}
    \caption{Distribution of error ratio across 100 incorrect generation results for Qwen2.5-7B-Instruct.}
    \label{fig:value_distribution}
    \vspace{-4mm}
\end{wrapfigure}

The automated optimization modeling task involves generating mathematical formulations that typically consist of $<$\textsc{variables}, \textsc{constraints}, \textsc{objectives}$>$. To investigate how and where errors occur, we conducted preliminary experiments using the base model Qwen-2.5-7B-Instruct~\citep{yang2025qwen3} on the ORLM training dataset~\citep{huang2025orlm}, which contains questions paired with their correct mathematical formulations. For each question, we compare the generated code against the ground-truth mathematical formulation using an LLM-as-a-judge approach to identify errors in the variables, constraints, and objectives.
We define the error ratio $\mathcal{E}$ of each instance as $\frac{N_{err\_var} + N_{err\_con}+ N_{err\_obj}}{N_{var} + N_{con} + N_{obj}}$, where $N(\cdot)$ is the number of the corresponding component.
As shown in Figure \ref{fig:value_distribution}, when errors occur, LLMs tend to introduce only a small fraction of errors rather than producing entirely incorrect formulations in most cases. The low average error ratio of 0.33 indicates that the variables, constraints, and objectives are relatively independent, thus limiting the error propagation. Additionally, we observed that certain types of errors are more likely to occur in specific components of the formulation. For instance, when modeling variables, LLMs often struggle to determine the appropriate data type (e.g., integer or continuous). As shown in Figure~\ref{fig:optimization_model_math_diff}, we illustrate the difference in error propagation between a general mathematical reasoning question and an optimization modeling question. This observation motivates us to systematically collect the most frequent error types from existing datasets and then synthesize new data that explicitly incorporates these common error patterns.

\section{Methodology}
\label{sec:methodology}

    \subsection{MIND: Error-driven Reverse Data Synthesis Pipeline}
\begin{figure}[tbp]
            \centering
            \includegraphics[width=1.0\linewidth]{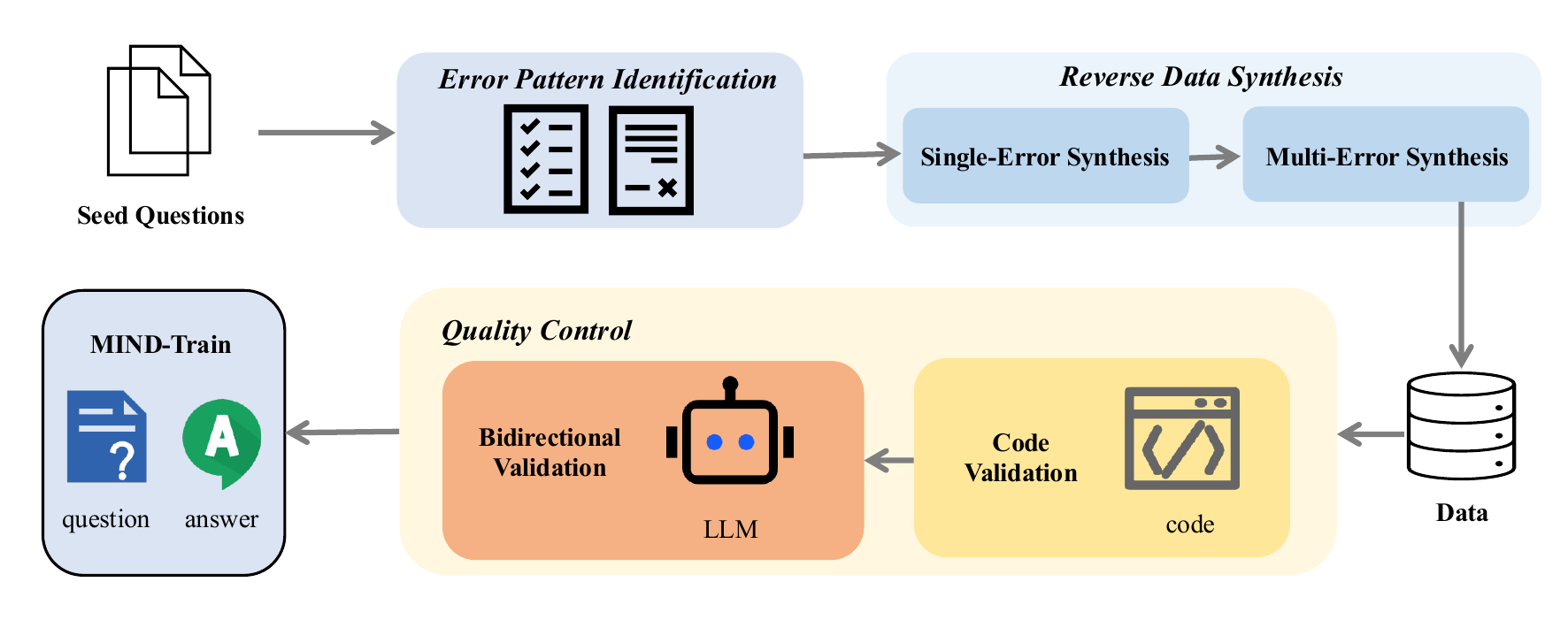}
            \caption{Overview of our proposed data synthesis pipeline.}
            \label{fig:data_synthesis_overview}
            \vspace{-1em}
        \end{figure}

        Motivated by our observations, we propose an error-driven reverse data synthesis pipeline, as illustrated in Figure~\ref{fig:data_synthesis_overview}. Our data generation process differs from prior work~\citep{huang2024mamo,yang2024optibench} in two key aspects: (1) we skip the costly collection of high-quality seed data by directly leveraging existing optimization modeling datasets as seeds; and (2) we deliberately target common error patterns that LLMs are prone to, thereby producing synthesized data that is inherently more challenging and better suited for robust model training. Our synthesis pipeline consists of three stages, including error pattern identification, reverse data synthesis, and quality control.

    \paragraph{Error Pattern Identification}
        Since our pipeline requires LLMs to make errors on the problems, we sample seed data from existing optimization modeling training datasets, namely OR-Instruct-Data-3K~\citep{huang2025orlm} and OptMATH-Train~\citep{lu2025optmath}. We then apply our base model to perform the reasoning process on this seed data and extract error patterns by comparing the generated code with the corresponding ground-truth formulations. The error pattern identification and extraction are accomplished by powerful LLMs such as DeepSeek-R1~\citep{guo2025deepseek}.

\begin{figure}[tbp]
            \centering
            \includegraphics[width=1.0\linewidth]{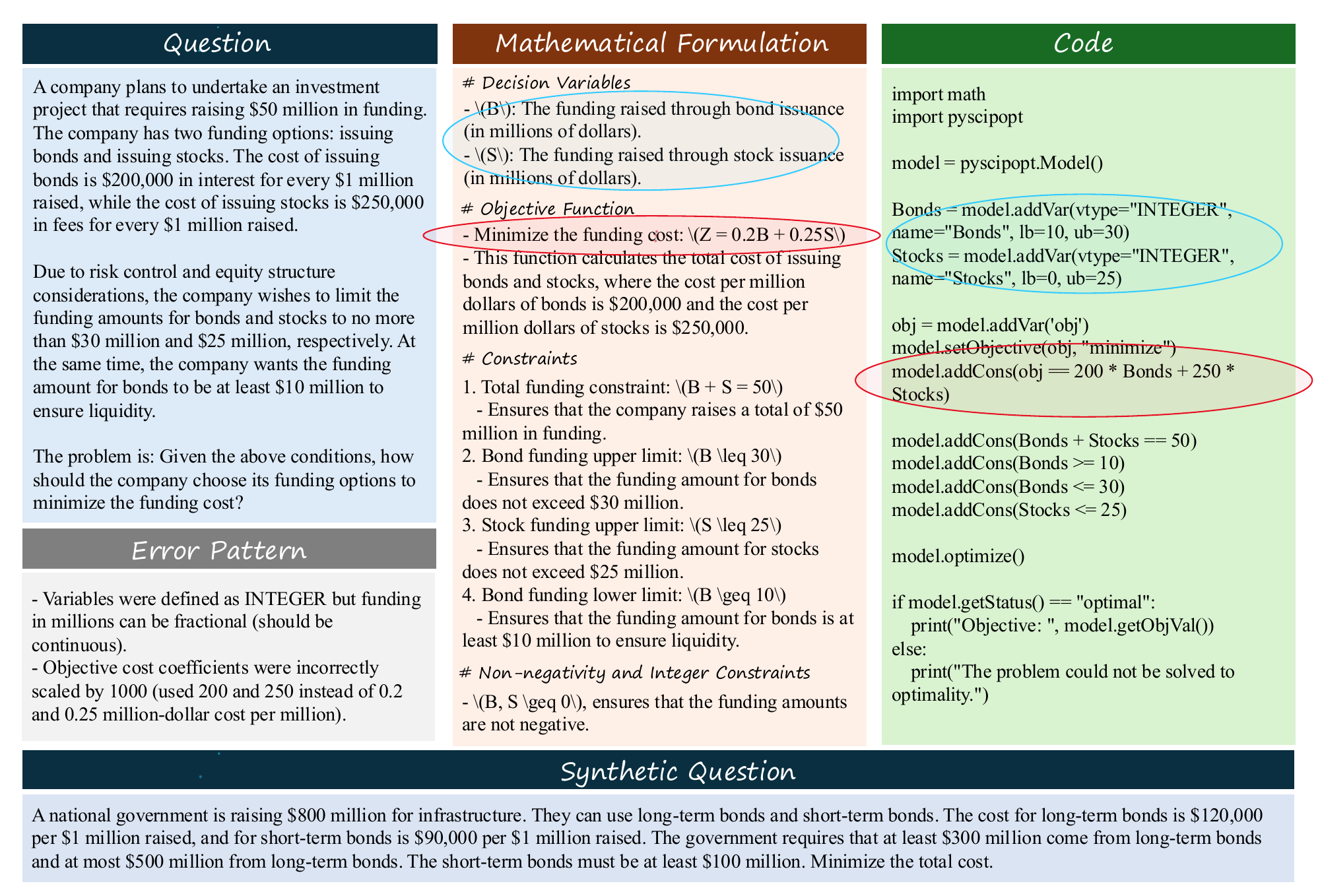}
            \caption{Example on single-error reverse data synthesis.}
            \label{fig:single_example}
            \vspace{-1em}
        \end{figure}

    \paragraph{Reverse Data Synthesis}
        After identifying the error patterns, we evolve the original questions into new ones by systematically incorporating these patterns. Since each question may contain multiple error types, we design two complementary strategies: single-error reverse data synthesis, where the LLM is instructed to focus on a single error pattern and generate a new problem that deliberately embeds a trap at that specific point (See example in Figure~\ref{fig:single_example}); and multi-error reverse data synthesis, which seeks to construct more challenging problems containing multiple potential error-prone points (See example in Figure~\ref{fig:multi_example}). Notably, LLMs are instructed to output not only the new problem but also its corresponding modeling solution.

    \paragraph{Quality Control}

        To ensure the quality of the generated data, we implement a two-stage quality control process:
        (1) \textit{Code validation}: We employ the target solver to verify the executability of the generated code and retain only those instances that can be successfully executed and solved to yield a reasonable solution (e.g., non-zero optimal value).
        (2) \textit{Bidirectional validation}: Since both the problem and its solution are evolved from the original question, we further employ another powerful LLM to directly solve the newly synthesized problem and compare the obtained optimal value against the ground-truth value in the synthesized dataset. Only the instances that pass this bidirectional validation are retained.

    We highlight that our reverse data synthesis method can leverage error patterns from different training datasets or industry scenario problems to generate diverse and challenging data. This approach significantly reduces the reliance on costly expert annotations for seed data, thereby improving both scalability and practicality.

\subsection{MIND: Error-driven Post-training Method}

    Existing approaches such as DAPO \citep{yu2025dapo}, SIRL \citep{chen2025solver} and GHPO \citep{liu2025ghpo} have sought to address the sparse reward problem on difficult samples through techniques like dynamic sampling, curated reward design, and adaptive prompt guidance. However, we argue that these methods still suffer from critical limitations, including insufficient guidance and distribution shifting on the policy model. To mitigate these challenges, we propose a novel framework termed \textbf{D}ynamic Supervised \textbf{F}ine-Tuning \textbf{P}olicy \textbf{O}ptimization (DFPO).

    \paragraph{Reward Design}
 
    We define modeling fidelity as the extent to which a mathematical formulation accurately represents the optimization problem it is intended to model. It is measured as the distance between the predicted formulation and the correct formulation, denoted by $\mathcal{E}$ (see Section~\ref{sec:motivation} for details). Objective accuracy represents the distance between the formulation's objective value and the ground-truth objective value. We hypothesize that, in general, higher fidelity in the mathematical formulation of an optimization problem is associated with more accurate objective values. Let $\mathcal{MF}_\theta$ denote the predicted mathematical formulation based on parameters $\theta$, and let $ \mathcal{MF}^*$ denote the one correct mathematical formulation. We introduce a modeling error measure, $\mathcal{E}(\mathcal{MF}_\theta, \mathcal{MF}^*)$, which captures discrepancies in variables, constraints, and the objective function. A larger $\mathcal{E}$ indicates greater deviation from the correct mathematical formulation. Our working hypothesis is that optimization modeling error and objective deviation are positively correlated in expectation. Formally, for two predicted problems $\mathcal{MF}_\theta^{(1)}$ and $\mathcal{MF}_\theta^{(2)}$, we generally expect:

    \begin{align}
        & \text{if } 
           \mathcal{E}\!\left(\mathcal{MF}_\theta^{(1)}, \mathcal{MF}^*\right) 
           < \mathcal{E}\!\left(\mathcal{MF}_\theta^{(2)}, \mathcal{MF}^*\right), \nonumber \\
        & \text{then } 
           \mathbb{E}\!\left[
           \left| \mathrm{Obj}\!\left(\mathcal{MF}_\theta^{(1)} \right) 
                  - \mathrm{Obj}\!\left( \mathcal{MF}^* \right) \right|
           \right]
           \;\leq\;
           \mathbb{E}\!\left[
           \left| \mathrm{Obj}\!\left(\mathcal{MF}_\theta^{(2)} \right) 
                  - \mathrm{Obj}\!\left( \mathcal{MF}^* \right) \right|
           \right],
    \end{align}
    
    where $\mathrm{Obj}(\mathcal{MF})$ denotes the objective value of the mathematical formulation $\mathcal{MF}$. This assumption underpins our reward design: by rewarding the agent based on the degree of modeling errors, we enable it to perceive the extent of such errors, thereby guiding the solutions to be structurally closer to the correct mathematical formulation.

    Therefore, we present our reward function as follows:   
    \begin{align*}
        R = \alpha \cdot R_{fid} + (1 - \alpha) \cdot R_{acc},
    \end{align*}
    where $\alpha = 0.2$, the modeling fidelity reward is defined as $R_{\text{fid}}=1-\frac{|obj_{\texttt{MIND}}-obj_{\texttt{GT}}|}{max(|obj_{\texttt{MIND}}|,|obj_{\texttt{GT}}|)}$, and the accuracy reward as
    \[
    R_{\text{acc}}=
    \begin{cases}
    1, & \text{if the answer is right},\\
    0, & \text{otherwise}.
    \end{cases}
    \]

    In this way, we mitigate the sparse reward problem by introducing a fidelity score, which provides partial credit when the generated mathematical formulation is close to, but not exactly identical to, the ground truth—a situation that accounts for the majority of cases.

    \begin{figure}
        \centering
        \includegraphics[width=1.0\linewidth]{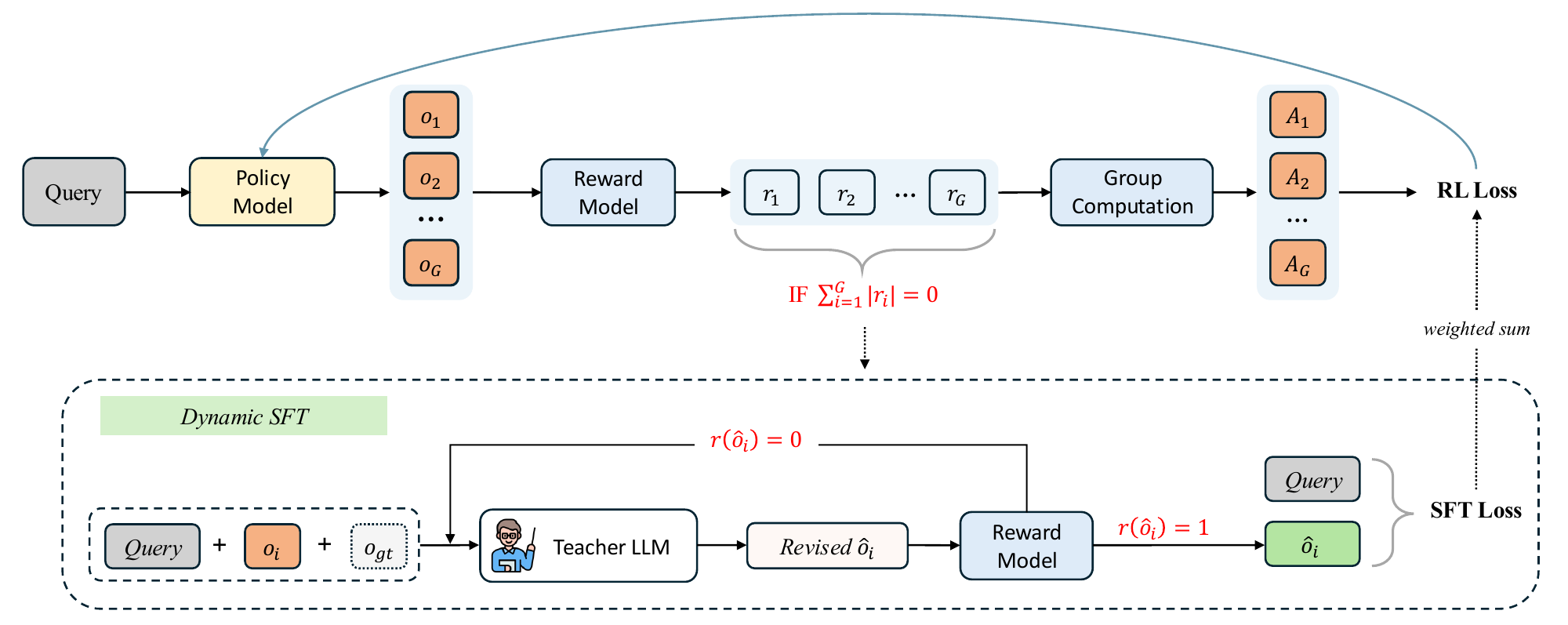}
        \caption{Overview of our proposed post-training method.}
        \label{fig:training_overview}
    \end{figure}

    \paragraph{Dynamic Supervised Fine-Tuning Policy Optimization}

        Standard GRPO and DAPO algorithms suffer from the sparse reward problem when dealing with difficult tasks, as they either perform inefficient explorations or discard unsuccessful samples. A straightforward remedy is to replace an incorrect rollout with the ground-truth solution or to provide partial solutions as hints when all rollouts fail, thereby alleviating the sparse reward issue. However, we contend that this approach still faces notable limitations: (1) not all ground-truth labels include the intermediate reasoning process, which is essential for fostering reasoning capabilities; and (2) the ground-truth solutions do not always align with the output distribution of the current policy model, making it difficult for the model to directly imitate the labeled behavior. We refer to this challenge as \textit{distributional shifting}.
        To overcome these limitations, we introduce Dynamic Supervised Fine-Tuning Policy Optimization (DFPO). Unlike existing methods, as illustrated in Figure \ref{fig:training_overview}, our approach leverages a stronger teacher LLM (e.g., DeepSeek-V3~\citep{liu2024deepseek}) to refine the base model’s incorrect responses, thereby ensuring that the corrected outputs remain closely aligned with the response distribution of the base model (see examples in Appendix~\ref{appendix:dynamic_sft_examples}). To enhance the reliability of this correction process, we provide the teacher LLM with access to the ground-truth solution. Once the teacher LLM generates a corrected response that is both accurate and distributionally consistent with the original incorrect rollout, this response is incorporated into the training process by computing the SFT loss. Finally, both the standard RL loss and the SFT loss are jointly utilized to guide the optimization of the policy model.

        \begin{align}
            \mathcal{L}_{\text{RL}}(\theta) = -\mathbb{E}_{\substack{ (q,a^*)\sim\mathcal{D}, \{o_i\}_{i=1}^G \sim \pi_{\theta_{\text{old}}}(\cdot|q) }} \Bigg[ \frac{1}{\sum_{i=1}^G |o_i|} \sum_{i=1}^G \sum_{t=1}^{|o_i|} \operatorname{min}\Big( & r_{i,t}(\theta)\hat{A}_{i,t}, \notag \\
            \operatorname{clip}\big(r_{i,t}(\theta), 1 - \varepsilon_{\text{low}}, 1 + \varepsilon_{\text{high}}\big) \hat{A}_{i,t} \Big) \Bigg]
        \end{align}
        $$
        \text{s.t. } 0 < |\{o_i | \text{is\_equivalent}(a^*, \mathrm{BS}(o_i))\}| < \gamma \times G,
        $$
        where
        \begin{align}
            r_{i,t}(\theta) = \frac{\pi_{\theta}(o_{i,t} \mid q, o_{i,<t})}{\pi_{\theta_{\text{old}}}(o_{i,t} \mid q, o_{i,<t})}, \quad \hat{A}_{i,t} = \frac{R_i - \text{mean}(\{R_i\}_{i=1}^G)}{\text{std}(\{R_i\}_{i=1}^G)}. 
        \end{align}

        \begin{equation}
            \mathcal{L}_{\text{NLL}}(\theta) = - \mathbb{E}_{\substack{ (q,a^*)\sim\mathcal{D}, \{{o_i\}_{i=1}^G \sim \pi_{\theta_{\text{old}}}(\cdot|q)}, \hat{o}_i \sim \pi_{\text{teacher}}(\cdot|q,\{o_i\}_{i=1}^G,o_{\text{gt}}) }} \Bigg[  \sum_{t=1}^{|\hat{o}_i|} \log \pi_\theta (a_t \mid s_t)\Bigg]
        \end{equation}

        $$
        \text{s.t. }  |\{o_i | \text{is\_equivalent}(a^*, \mathrm{BS}(o_i))\}| = 0.
        $$

        \begin{equation}
            \mathcal{L}_{\text{DFPO}}(\theta) = \mathcal{L}_{\text{RL}}(\theta) + \beta \cdot \sqrt{\frac{n_{\texttt{SFT}}}{n_{\texttt{RL}}}} \cdot \mathcal{L}_{\text{NLL}}(\theta),
        \end{equation}

        where $n_{\texttt{SFT}}$ and $n_{\texttt{RL}}$ denote the numbers of SFT and RL responses in each training batch.

\section{Experiments}
\label{sec:exp}
We conduct extensive experiments to study the effectiveness of our proposed MIND on automated optimization modeling. We aim to study the following research questions (RQ):

    \begin{enumerate}[label=\textbf{RQ\arabic*}, leftmargin=3em]
        \item Can MIND improve the base model’s performance in automated optimization modeling? 
        \item How does MIND compare with state-of-the-art automated optimization methods? 
        \item How effective is the proposed error-driven reverse data synthesis pipeline? 
        \item How effective is the proposed error-driven DFPO post-training method? 
        \item Can MIND generalize to out-of-distribution automated optimization modeling scenarios? 
    \end{enumerate}

\subsection{Experimental Setup}

    Following existing work \citep{huang2025orlm, chen2025solver}, Qwen-2.5-7B-Instruct is employed as our base model. To further align with recent advances, we also adopt Qwen3-8B, a newly released and widely adopted open-source model, as an additional base model. We construct the MIND-Train dataset by synthesizing data from the seed datasets OR-Instruct-Data-3K~\citep{huang2025orlm} and OptMATH-Train \citep{lu2025optmath}. We note that Qwen2.5-7B-Instruct is specifically used for the error pattern identification stage in the data synthesis pipeline. A detailed summary of MIND-Train is provided in Appendix~\ref{appendix:mind_train}. Finally, we sample 10,000 instances for training.

    \paragraph{Benchmarks \& Baselines} 

        We conduct comprehensive evaluations on NL4Opt~\citep{ramamonjison2023nl4opt}, IndustryOR~\citep{huang2025orlm}, MAMO~\citep{huang2024mamo} (EasyLP and ComplexLP), OptMATH-Bench~\citep{lu2025optmath}, and OptiBench~\citep{yang2024optibench}. Further details on the benchmarks can be found in Appendix~\ref{appendix:benchmark}. We compare our method against GPT-4~\citep{achiam2023gpt}, OpenAI o3~\citep{jaech2024openai}, Deepseek-V3~\citep{liu2024deepseek}, Deepseek-R1~\citep{guo2025deepseek}, Qwen2.5-7B-Instruct~\citep{yang2025qwen3}, Qwen3-8B~\citep{yang2025qwen3}, Autoformulator~\citep{astorga2024autoformulation}, Chain-of-Experts~\citep{xiao2023chain}, Step-Opt~\citep{wu2025step}, OptiMUS~\citep{ahmaditeshnizi2023optimus}, ORLM~\citep{huang2025orlm}, LLMOPT~\citep{jiang2024llmopt}, OptMATH~\citep{lu2025optmath}, and SIRL~\citep{chen2025solver}.

    \paragraph{Evaluation and Metrics}

        Following previous work, we evaluate all methods with pass@1 accuracy in a zero-shot setting. A  solution is deemed correct if the relative error between the objective value produced by the LLM and the ground-truth objective value is less than $10^{-6}$.

\subsection{Main Results}

    \textit{\textbf{RQ1}: MIND consistently improves automated modeling performance}.
    As shown in Table~\ref{tab:main_table}, MIND-Qwen2.5-7B enhances the base model’s automated modeling performance by approximately 14.3\% across six benchmarks.    
    On relatively simpler benchmarks such as NL4Opt and EasyLP, MIND yields moderate improvements over already strong baseline scores. In contrast, on more challenging benchmarks such as IndustryOR, ComplexLP, and OptMATH, the performance gains are significant, with an average improvement of 24.1\%. Moreover, we observe that on OptiBench, which primarily consists of tabular data, MIND-Qwen2.5-7B achieves only marginal improvement, likely due to the limited representation of similar problem types in the training dataset. Furthermore, MIND-Qwen3-8B enhances its base model's performance by approximately 31.0\%, providing additional evidence that MIND is effective across different base model architectures.

\begin{table}[htbp!]
  \caption{Performance comparison of models on benchmarks (pass@1$\uparrow$). Methods marked with * indicate that their results are taken from the original or reproduced papers.}
  \setlength{\tabcolsep}{0.8mm} 
  \label{tab:main_table}
  \centering
  \begin{threeparttable}
  \resizebox{\textwidth}{!}{
    \begin{tabular}{l|lccccccc}
    \toprule
    \toprule
    \textbf{Category} & \textbf{Methods} & \textbf{NL4Opt} & \textbf{IndustryOR} & \textbf{EasyLP} & \textbf{ComplexLP} & \textbf{OptMATH} & \textbf{OptiBench} & \textbf{Macro AVG} \\
    \midrule
    \midrule
    \multirow{2}{*}{\textbf{Proprietary}} 
    & GPT-4* & 89.0\% & 33.0\% & 87.3\% & 49.3\% & 16.6\% & 68.6\% & 57.4\% \\
    & OpenAI o3* & 69.4\% & 44.0\% & 77.1\% & 51.2\% & 44.0\% & 58.6\% & 57.4\% \\
    \midrule
    \multirow{4}{*}{\textbf{Open-Source}} & Deepseek-V3* & 95.9\% & 37.0\% & 88.3\% & 50.2\% & 44.0\% & 71.6\% & 64.5\% \\
    & Deepseek-R1* & 82.4\% & 45.0\% & 87.2\% & 67.9\% & 40.4\%  & 66.4\% & 61.9\%  \\
    & Qwen2.5-7B-Instruct & 89.0\% & 24.0\% & 89.4\% & 31.5\% & 3.0\%  & 53.2\% & 48.4\% \\
    & Qwen3-8B & 72.2\% & 14.0\% & 76.8\% & 17.2\% & 7.2\%  & 36.5\% & 37.3\% \\
    \midrule
    \multirow{3}{*}{\textbf{TTS-based}}
    & Autoformulator* & 92.6\% & 48.0\% & - & 62.3\% & -  & - & - \\
    & Chain-of-Experts* & 64.2\% & - & - & 40.2\% & -  & - & - \\
    & OptiMUS* & 78.8\% & 31.0\% & 77.0\% & 43.6\% & 20.2\% &  45.8\% &  49.4\% \\
    \midrule
    \multirow{5}{*}{\textbf{Fine-Tuning}} & ORLM-Llama3-8B* & 85.7\% & 24.0\% & 82.3\% & 37.4\% & 2.6\%  &  51.1\% & 47.2\% \\
    & Step-Opt-Llama3-8B* & 84.5\% & 36.4\% & 85.3\% & 61.6\% & -  &  - & - \\
    & LLMOPT-Qwen2.5-14B* & 80.3\% & 29.0\% & 89.5\% & 44.1\% & 12.5\%  & 53.8\% & 51.1\% \\
    & OptMATH-Qwen2.5-7B* & 94.7\% & 20.0\% & 86.5\% & 51.2\% & 24.4\%  & 57.9\% & 55.8\% \\
    & OptMATH-Qwen2.5-32B* & 95.9\% & 31.0\% & 89.9\% & 54.1\% & 34.7\%  &  66.1\% & 62.0\% \\
    \midrule
    \multirow{2}{*}{\textbf{RLVR}} & SIRL-Qwen2.5-7B* & 96.3\% & 33.0\% & 91.7\% & 51.7\% & 30.5\%  & 58.0\% & 60.2\% \\
    & SIRL-Qwen2.5-32B* & 98.0\% & 42.0\% & 94.6\% & 61.1\% & 45.8\%  & 67.4\% & 68.2\% \\
    \midrule
    \multirow{2}{*}{\textbf{Ours}} 
    & MIND-Qwen2.5-7B & 96.7\% & 34.0\% & 92.2\% & 60.1\% & 36.7\% & 56.7\% & 62.7\% \\
    & MIND-Qwen3-8B & 95.1\% & 42.0\% & 92.7\% & 76.8\% & 41.0\%  & 62.0\% & 68.3\% \\
    \bottomrule
    \bottomrule
    \end{tabular}
  }
\end{threeparttable}
\end{table}

    \textit{\textbf{RQ2}: MIND outperforms state-of-the-art automated modeling methods}. We compare MIND-Qwen2.5-7B and MIND-Qwen3-8B against a range of representative approaches, including proprietary models, agent-based frameworks, and training-based methods. As reported in Table~\ref{tab:main_table}, MIND-Qwen2.5-7B achieves superior average performance compared with all baseline models of comparable parameter size. In particular, relative to prior training-based approaches, MIND-Qwen2.5-7B demonstrates remarkable improvements on more challenging benchmarks such as IndustryOR, ComplexLP, and OptMATH. These results highlight the effectiveness of our reverse data synthesis pipeline and our proposed DFPO method. Furthermore, we observe that MIND-Qwen3-8B achieves competitive performance across all baselines, including larger models such as Deepseek-V3, GPT-4, OptMATH-Qwen2.5-32B, and SIRL-Qwen2.5-32B.

\subsection{Ablation Study}

    \paragraph{Data Synthesis Framework (\textit{RQ3})}

\begin{wrapfigure}{r}{0.50\textwidth}
            \vspace{-1mm}
            \centering
            \includegraphics[width=0.49\textwidth]{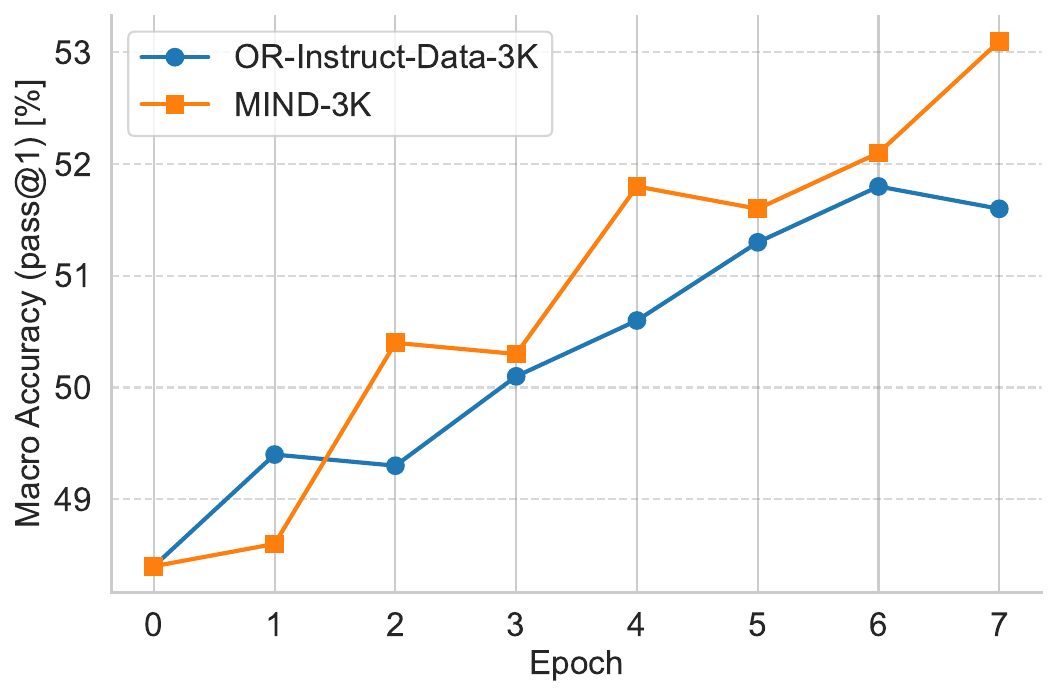}
            \vspace{-2.5mm}
            \caption{Ablation study of data synthesis methods across six benchmarks.}
            \label{fig:data_synthesis_analysis}
            \vspace{-6mm}
        \end{wrapfigure}

        To assess the effectiveness of our data synthesis approach, we employ DAPO~\citep{yu2025dapo} to train Qwen-2.5-7B-Instruct from scratch on two datasets: OR-Instruct-Data-3K (3,000 instances) and MIND-3K (3,000 instances), where MIND-3K is generated from OR-Instruct-Data-3K using our proposed reverse data synthesis technique. As illustrated in Figure\ref{fig:data_synthesis_analysis}, the model trained on MIND-3K achieves consistently higher accuracy gains as training progresses, indicating that our error-driven reverse data synthesis method yields superior sample efficiency. Furthermore, Table~\ref{tab:data_synthesis_ablation_study_table} reports a detailed performance comparison after seven training epochs across six benchmarks, showing that the model trained on MIND-3K outperforms its counterpart trained on the majority of benchmarks. (See the details of the ablation study on single-error and multi-error strategies in Appendix~\ref{appendix:decomposition_ablation_study}).

        \begin{table}[htbp!]
    \centering
    \caption{Ablation results of the data synthesis pipeline on Qwen2.5-7B-Instruct (pass@1$\uparrow$).}
    \label{tab:data_synthesis_ablation_study_table}
    \resizebox{1.0\textwidth}{!}{
        \begin{tabular}{cccccccc}
        \toprule
        \textbf{Data} & \textbf{NL4OPT} & \textbf{IndustryOR} & \textbf{EasyLP} & \textbf{ComplexLP}  & \textbf{OptMATH} & \textbf{OptiBench} & \textbf{Macro AVG} \\
        \midrule
        OR-Instruct-Data-3K & 93.9\% & 25.0\% & 90.7\% & 35.0\% & 10.8\% & 54.4\% & 51.6\% \\
        MIND-3K & 94.3\% & 30.0\% & 90.8\% & 39.9\% & 7.8\% & 55.5\% & 53.1\% \\
        \bottomrule
        \end{tabular}
    }
\end{table}

    \paragraph{Post-Training Framework (\textit{RQ4})}

        To verify the effectiveness of DFPO, we compare it with DAPO, SFT, and SFT+GRPO under our reward design. Both methods are trained on the same dataset of 10,000 instances (Table~\ref{tab:main_table}) and use the same chain-of-thought prompt. As shown in Table~\ref{tab:post_training_ablation_study_table}, DFPO outperforms DAPO by about 1.9\% in macro-average accuracy across six benchmarks, with a notable gain of 10.2\% on OptMATH. This demonstrates that DFPO provides more effective learning signals for difficult problems. We highlight that while DAPO receives sufficient learning signals from easy problems through reinforcement learning, it receives limited signals from difficult problems. In contrast, DFPO leverages dynamic SFT techniques to capture additional learning signals from difficult problems, as evidenced by its improvement over DAPO in OptMATH. Furthermore, we observe that applying SFT alone on a relatively small training dataset (10,000 instances) does not yield significant performance gains. However, when used as a warm start for GRPO, the model achieves notable improvement, though it still lags behind DFPO on challenging benchmarks.

        \begin{table}[htbp!]
    \centering
    \caption{Ablation results for post-training method on Qwen2.5-7B-Instruct (pass@1$\uparrow$).}
    \label{tab:post_training_ablation_study_table}
    \resizebox{0.90\textwidth}{!}{
        \begin{tabular}{ccccccccc}
        \toprule
        \textbf{Methods} & \textbf{NL4OPT} & \textbf{IndustryOR} & \textbf{EasyLP} & \textbf{ComplexLP}  & \textbf{OptMATH} & \textbf{OptiBench} & \textbf{Macro AVG} \\
        \midrule
        DFPO & 96.7\% & 34.0\% & 92.2\% & 60.1\% & 36.7\% & 56.7\% & 62.7\% \\
        DAPO & 96.7\% & 33.0\% & 92.5\% & 58.6\% & 26.5\% & 57.5\% & 60.8\% \\
        SFT & 92.2\% & 31.0\% & 85.4\% & 37.4\% & 9.6\% & 55.9\% & 51.9\% \\
        SFT+GRPO & 93.9\% & 34.0\% & 90.2\% & 54.7\% & 28.3\% & 57.0\% & 59.7\% \\
        \bottomrule
        \end{tabular}
    }
\end{table}

\subsection{Generalization Study (\textit{RQ5})}

    In this paper, we introduce MIND-Bench, a benchmark that comprises 69 carefully curated operations research problems drawn from industry scenarios and textbooks (see Appendix~\ref{appendix:mind_bench} for details). As shown in Table~\ref{tab:generalization_study}, MIND-Qwen2.5-7B demonstrates superior generalization on MIND-Bench compared with the state-of-the-art post-training model SIRL-Qwen2.5-7B, although it still lags behind the 671B-parameter foundation models Deepseek-V3 and Deepseek-R1. Furthermore, MIND-Qwen3-8B is competitive with Deepseek-V3, Deepseek-R1, and SIRL-Qwen2.5-32B, all of which have larger parameter sizes.

    \begin{table}[htbp!]
    \centering
    \caption{Performance comparison of our proposed MIND and baselines on \textsc{MIND-Bench}.}
    \label{tab:generalization_study}
    \resizebox{0.8\textwidth}{!}{
        \begin{tabular}{cccccccc}
        \toprule
        \textbf{Deepseek-V3} & \textbf{Deepseek-R1} & \textbf{Qwen2.5-7B-Instruct} & \textbf{Qwen3-8B} \\
        \midrule
        66.7\% & 75.4\% & 29.0\% & 27.5\%  \\
        \midrule
        \textbf{SIRL-Qwen2.5-7B} & \textbf{SIRL-Qwen2.5-32B} & \textbf{MIND-Qwen2.5-7B} & \textbf{MIND-Qwen3-8B} \\
        \midrule
        46.4\% & 65.2\% & 50.7\% & 68.1\% \\
        \bottomrule
        \end{tabular}
    }
\end{table}

\section{Conclusion}
\label{sec:conclusion}
In this paper, we empirically show that modeling errors are often localized within specific semantic segments. Motivated by this finding, we propose a novel error-driven learning framework, which customizes the whole model training framework from data synthesis to post-training. Our study highlights two key insights: (1) Data synthesis: Domain-specific LLM performance depends heavily on the diversity, quality, and quantity of training data. (2) Post-training: Due to the complexity of automated optimization modeling tasks, LLMs often struggle to receive sufficient learning signals through reinforcement learning alone on difficult problems. Together, these insights advance the understanding and development of LLMs for automated optimization modeling.

\bibliography{main}
\bibliographystyle{main}

\clearpage
\newpage
\appendix
\section*{APPENDIX}
\addcontentsline{toc}{section}{Appendix}
\startcontents[appendix]
\printcontents[appendix]{}{1}{}

\section{Dataset}
\label{appendix:dataset}

    \subsection{Benchmark Dataset}
\label{appendix:benchmark}

    \paragraph{NL4Opt~\citep{ramamonjison2023nl4opt}}

        NL4OPT contains 245 high-quality questions, validated by ~\citep{lu2025optmath}. It includes only linear programming (LP) problems across various contexts. As the first curated dataset derived from the NL4OPT Competition, NL4OPT is considered an easy benchmark, featuring simple constraints and scenarios.

    \paragraph{MAMO~\citep{huang2024mamo}}

        MAMO consist of 642 high-quality questions in EasyLP and 203 high-quality questions in ComplexLP, as revised by~\citep{chen2025solver}. It focuses on linear programming (LP) and mixed-integer linear programming (MILP) problems. Compared with other benchmarks, MAMO primarily emphasizes LLM modeling skills on MILP, which constitutes the majority of real-world optimization problems.

    \paragraph{IndustryOR~\citep{huang2025orlm}}

        IndustryOR contains 100 questions collected from real-world optimization scenarios across various sectors, as verified by~\citep{chen2025solver}. It includes integer programming (IP), linear programming (LP), mixed-integer linear programming (MILP), and nonlinear programming (NLP), and other specialized formulations. Unlike other benchmarks, IndustryOR specifically targets industrial applications, capturing the complexity of real-world optimization problems.

    \paragraph{OptiBench~\citep{yang2024optibench}}

        OptiBench contains 605 questions collected from textbooks~\citep{bertsimas1997introduction, conforti2014integer, wolsey2020integer}. It includes integer programming (IP), linear programming (LP), mixed-integer linear programming (MILP), and nonlinear programming (NLP). Compared with other benchmarks, OptiBench features extensive tabular data, enabling the evaluation of LLMs’ ability to understand and reason with tables.

    \paragraph{OptMATH-Bench~\citep{lu2025optmath}}

        OptMATH-Bench contains 166 carefully curated questions constructed by human experts. It includes integer programming (IP), linear programming (LP), mixed-integer linear programming (MILP), nonlinear programming (NLP), and second-order cone programming (SOCP). Compared with other benchmarks, OptMATH-Bench features longer natural language contexts and more complex constraints, enabling the evaluation of LLMs’ long-context optimization modeling capacity.

    OptMATH~\citep{lu2025optmath} and SIRL~\citep{chen2025solver} highlight that portions of the problem statements in benchmarks contain ambiguities, making it difficult for both LLMs and human experts to determine whether a variable should be treated as integer or continuous, depending on the practical context. Following their approach, we also adopt a rule-based substitution method. We consider a case as passed if the optimal solution, whether derived under the integer or continuous assumption, matches the ground truth, i.e., the objective absolute difference between the LLM-generated mathematical formulation and the ground-truth formulation is less than $10^{-6}$.

\subsection{Seed Dataset}
\label{appendix:seed_dataset}

    \paragraph{OR-Instruct-Data-3K}

        OR-Instruct-Data-3K, released by ORLM, contains 3,000 training instances (a subset of the full 30,000 ORLM training examples), each including the \texttt{question}, \texttt{mathematical formulation}, and \texttt{code}.

    \paragraph{OptMATH-Train}

        OptMATH-Train, released by OptMATH, contains 200,000 training instances, each including the \texttt{question}, \texttt{mathematical formulation}, and \texttt{code}.

\subsection{MIND-Train Dataset}
\label{appendix:mind_train}

    \paragraph{MIND-Train Statistics}

\begin{figure}[htbp!]
            \centering
            \includegraphics[width=1.0\linewidth]{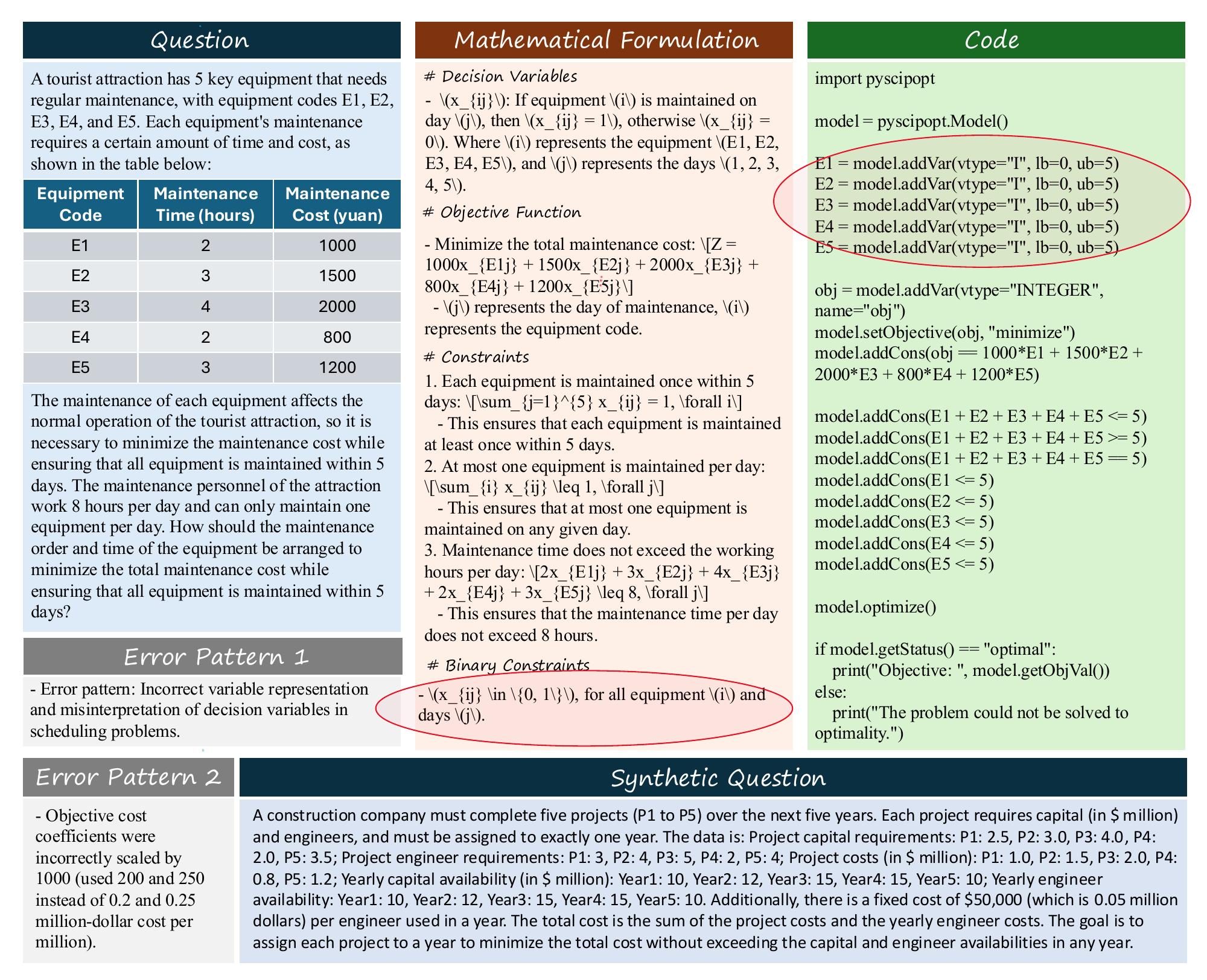}
            \caption{Example on multi-error reverse data synthesis.}
            \label{fig:multi_example}
        \end{figure}

        As shown in Table~\ref{tab:quality_control}, we provide statistical information for MIND-Train, summarizing the question examples across three stages of the reverse data synthesis pipeline. We present a multi-error reverse data synthesis example in Figure~\ref{fig:multi_example}, complementing the single-error reverse data synthesis example (see Figure~\ref{fig:single_example}). We note that error pattern 1 comes from Figure~\ref{fig:multi_example}, while error pattern 2 comes from Figure~\ref{fig:single_example}.

        \begin{figure}[htbp!]
            \centering
            \includegraphics[width=1.0\linewidth]{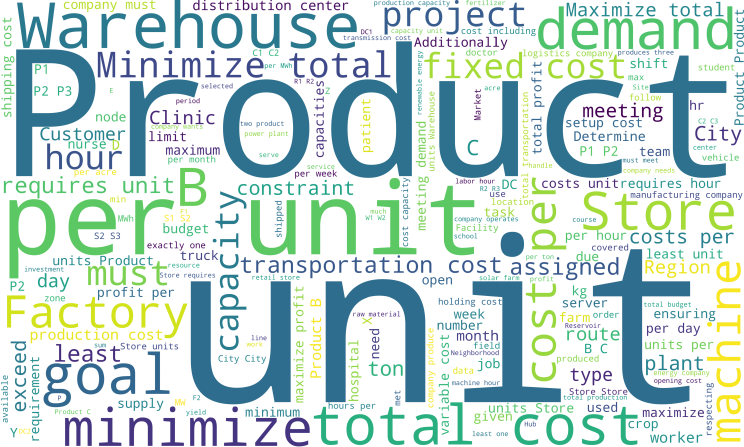}
            \caption{The statistical word cloud of MIND-Train.}
            \label{fig:word_cloud_analysis}
        \end{figure}

        \begin{table}[htbp!]
    \centering
    \caption{MIND-Train dataset construction summary. The single-error strategy uses DeepSeek-R1-0528, while the multi-error strategy uses DeepSeek-V3.1-Think.}
    \label{tab:quality_control}
    \resizebox{\textwidth}{!}{
    \begin{tabular}{cccccccc}
    \toprule
    \toprule
    \multirow{2}{*}{Synthesis Category} & \multirow{2}{*}{Seed Data} & \multirow{2}{*}{Initial Count} & \multicolumn{2}{c}{Code} & \multicolumn{2}{c}{Bidirectional} & \multirow{2}{*}{Passed Rate} \\
    \cmidrule(lr){4-5} \cmidrule(lr){6-7} 
    & & & Count & Rate & Count & Rate  \\
    \midrule
    \midrule
    Single-Error & ORLM & 5033 & 5016 & 99.66\% & 2007 & 40.01\% & 39.88\% \\
    Multi-Error & ORLM & 2977 & 2910 & 97.75\% & 1795 & 61.68\% & 60.30\% \\
    Single-Error & OptMATH & 9676 & 5950 & 61.49\% & 2961 & 49.76\% & 30.60\% \\
    Multi-Error & OptMATH & 2850 & 2102 & 73.75\% & 1494 & 71.07\% & 52.42\% \\
    Multi-Error & ALL & 2473 & 1843 & 74.52\% & 1406 & 76.29\% &  56.85\% \\
    \midrule
    Total & - & 23009 & 17821 & 77.45\% & 9663 & 54.22\%  & 42.00\% \\
    \bottomrule
    \bottomrule
    \end{tabular}
    }
\end{table}

    \paragraph{Word Cloud Analysis}

        As shown in Figure~\ref{fig:word_cloud_analysis}, the word cloud highlights diverse automatic optimization modeling topics (e.g. hospital, transportation, machine, warehouse, surgery, facility, energy, product).

    \paragraph{Gerund Pairs Analysis}

\begin{figure}[htbp!]
            \centering
            \includegraphics[width=1.0\linewidth]{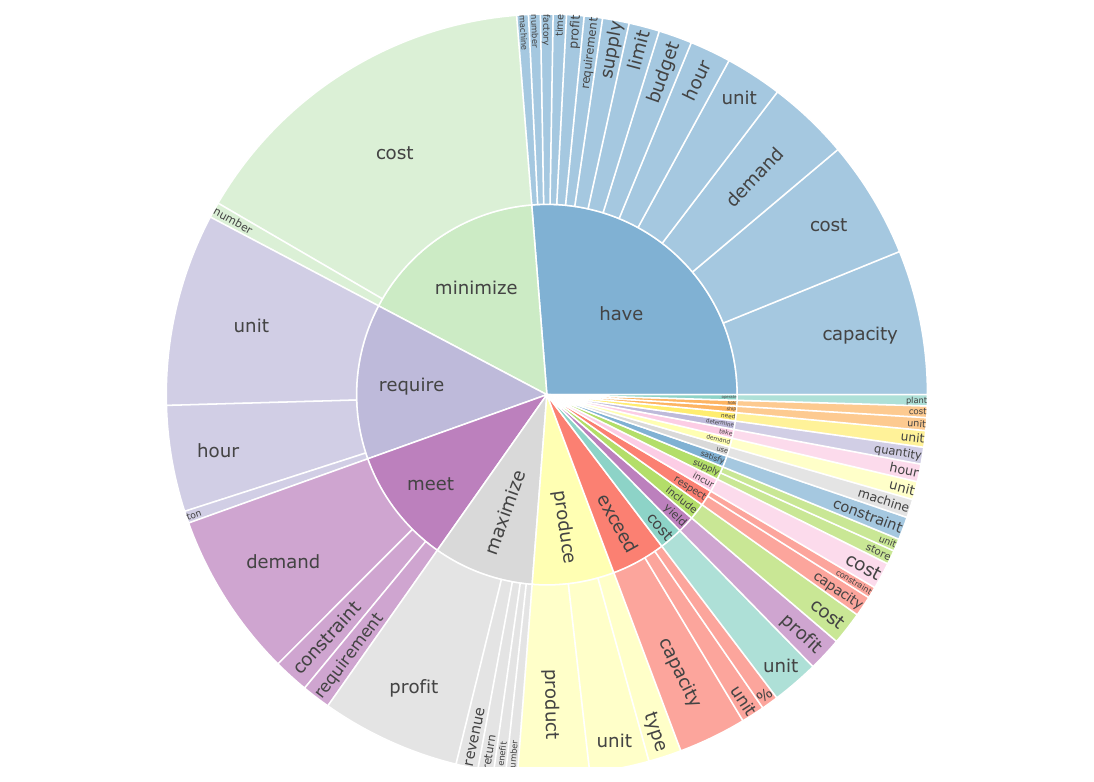}
            \caption{Top 50 gerund pairs of questions in MIND-Train.}
            \label{fig:gerund_pairs_analysis}
        \end{figure}

        As shown in Figure~\ref{fig:gerund_pairs_analysis}, we use en\_core\_web\_sm~\citep{spacy2} to extract gerund pairs. The top 50 frequent gerund pairs represent typical optimization modeling patterns.

    \paragraph{Length Distribution Analysis}

\begin{figure}[htbp!]
            \centering
            \begin{subfigure}[b]{0.78\linewidth}
                \centering
                \includegraphics[width=\linewidth]{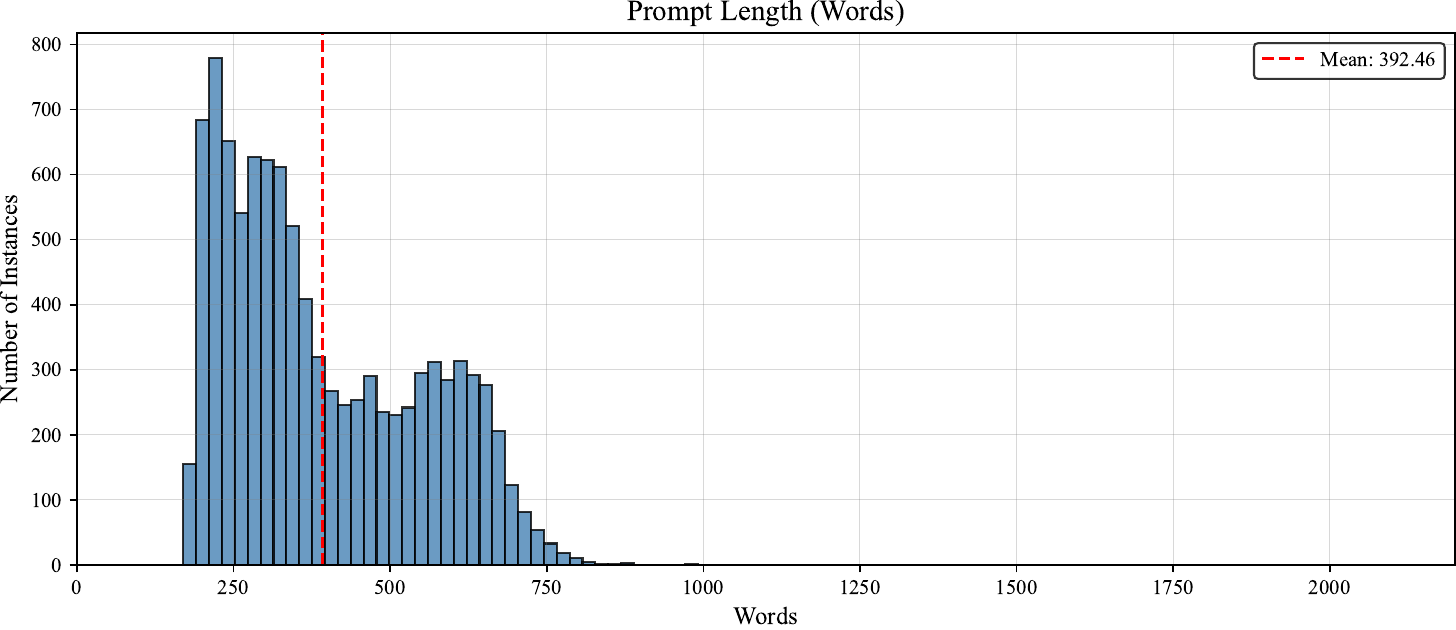}
            \end{subfigure}
            \vspace{4mm} 
            \begin{subfigure}[b]{0.8\linewidth}
                \centering
                \includegraphics[width=\linewidth]{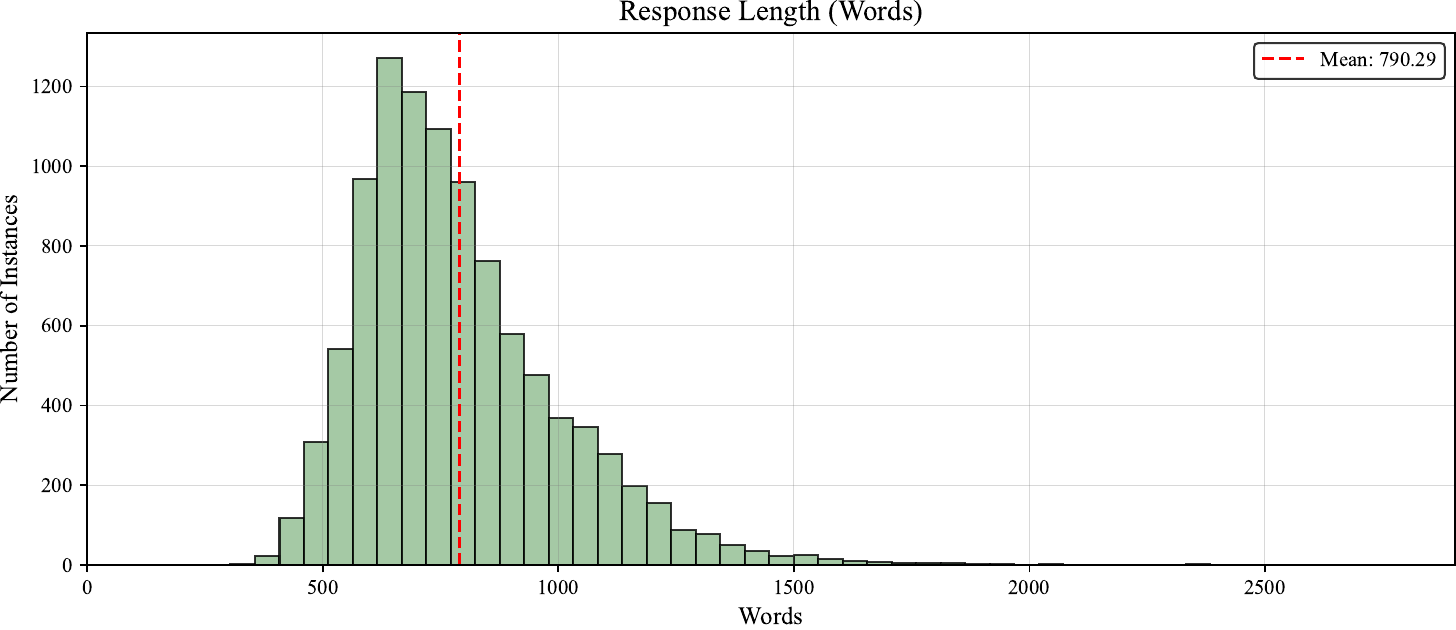}
            \end{subfigure}
            \caption{Length distribution of the training dataset for MIND-Qwen2.5-7B.}
            \label{fig:length_analysis}
        \end{figure}

        As shown in Figure~\ref{fig:length_analysis}, we examine the word length distributions of the prompts and responses in the training dataset (10,000 instances). The prompts exhibit an average length of 392 words, with most within the 200–600 word range. In comparison, responses are substantially longer, averaging 790 words, with the majority falling between 500 and 1,200 words.

    To increase the diversity of the training dataset, we sample 5,000 instances from MIND-Train, 1,000 instances from OR-Instruct-Data-3K, and 4,000 instances from OptMATH-Train. In total, we use 10,000 instances to train the Qwen2.5-7B-Instruct.

\subsection{MIND-Bench Dataset}
\label{appendix:mind_bench}

\begin{figure}[htbp!]
        \centering
        \includegraphics[width=1.0\linewidth]{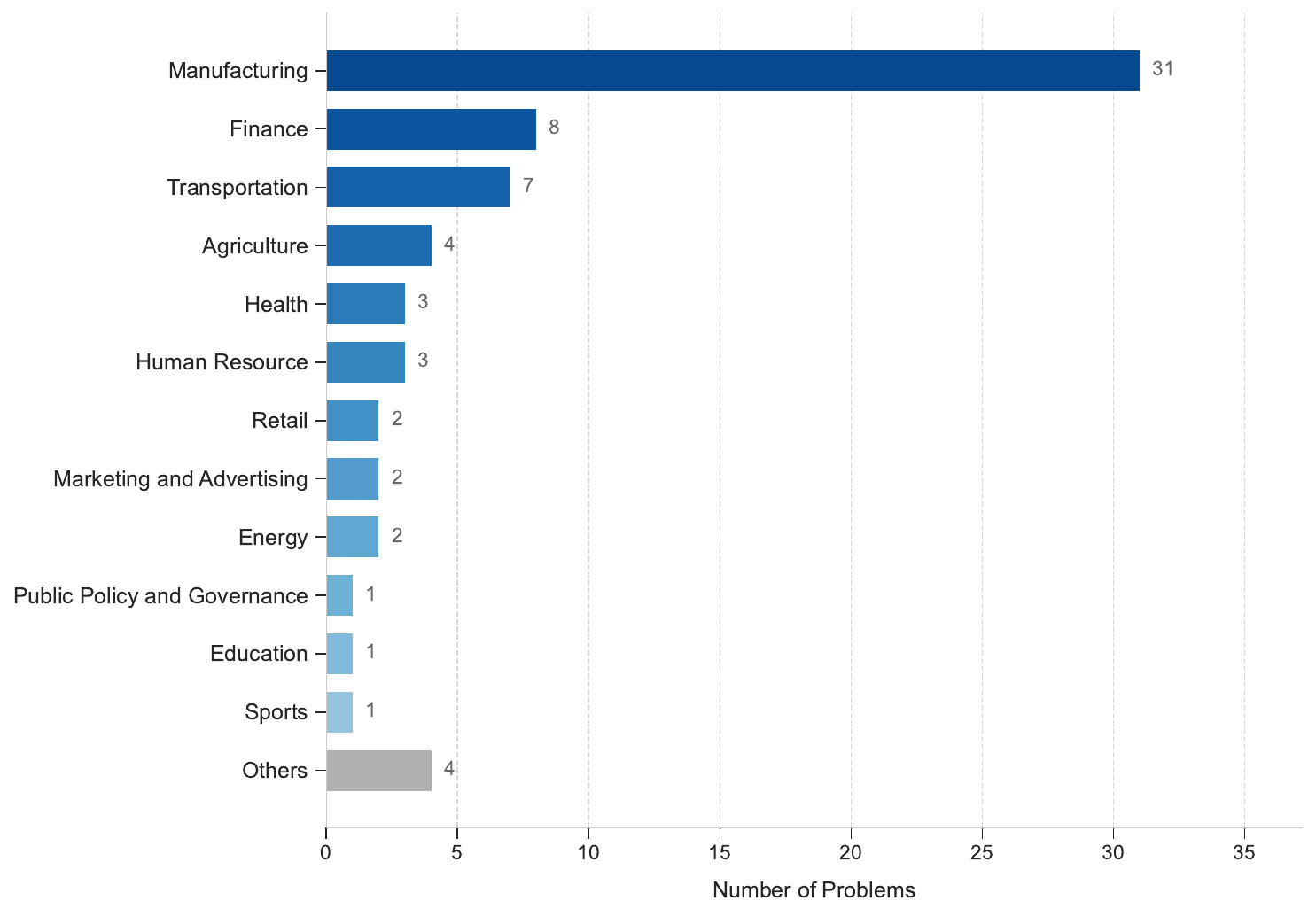}
        \caption{Scenario statistics of MIND-Bench.}
        \label{fig:mind_bench_scenarios}
    \end{figure}

\begin{figure}[htbp!]
        \centering
        \includegraphics[width=1.0\linewidth]{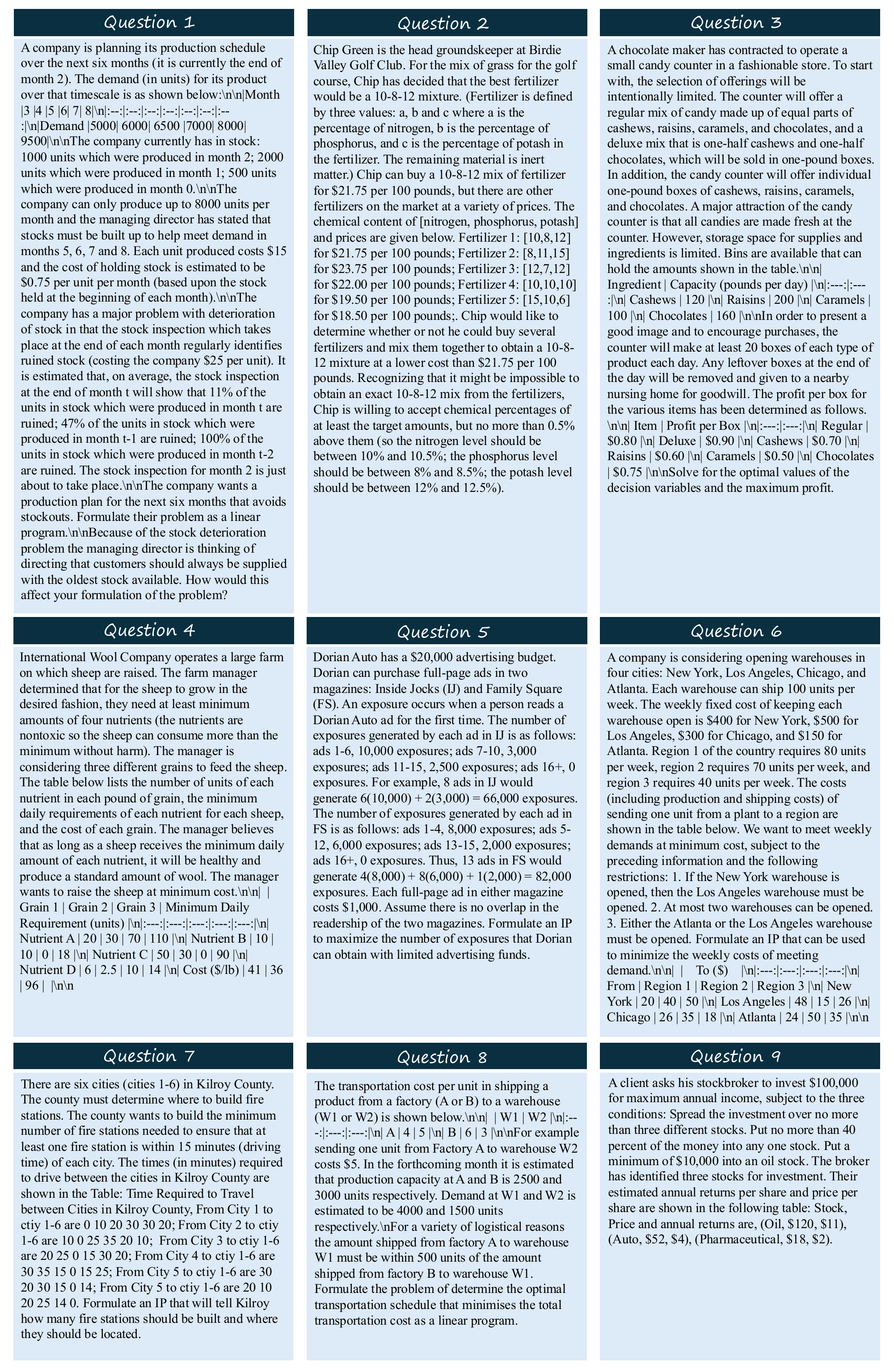}
        \caption{Problem examples from MIND-Bench.}
        \label{fig:mind_bench_examples}
    \end{figure}

    To evaluate the generalization ability of LLMs, we carefully curated 69 questions derived from textbooks or industry scenarios (See details in Figure~\ref{fig:mind_bench_scenarios}). These questions originate from out-of-distribution data sources that differ from those of other public benchmarks and training datasets. Examples of the questions are shown in Figure~\ref{fig:mind_bench_examples}. For questions in MIND-Bench, there is no ambiguity regarding variable types, and we do not use a rule-based substitution method.

\section{Prompt Templates}
\label{appendix:prompt}

    \subsection{Prompt Template for Preliminary Results}
    \label{appendix:motivation_prompt}
    
        \begin{tcolorbox}[
    enhanced, 
    colback=blue!5!white, 
    colframe=blue!50!white,
    coltitle=white,
    fonttitle=\bfseries,
    title=Prompt template used for preliminary results,
    arc=6pt, 
    boxrule=1pt,
    width=\textwidth,
    sharp corners=south,
    drop shadow
]
You will be given:

- A natural language description of an optimization problem.

- A correct mathematical formulation for the optimization problem.

- PySCIPOpt code that may contain errors for the optimization problem.

```

$\{\texttt{question}\}$

''' 

is the natural language description of an optimization problem.

```

$\{\texttt{mathematical formulation}\}$

''' 

is the correct mathematical formulation for the optimization problem.

```

$\{\texttt{python}\}$

```

is the PySCIPOpt code that may contain errors for the optimization problem.

We define a mathematical formulation size function $S(\cdot)$ as follows:

\begin{align}
    S(\mathcal{MF}) ={}& N_{\text{var}} + N_{\text{obj}} + N_{\text{cont}},
\end{align}

where $N_{\text{var}}$, $N_{\text{obj}}$, and $N_{\text{cont}}$ denote the numbers of variables, objectives (always set to 1), and constraints, respectively.

Your task is to analyze the consistency between the correct formulation and its implementation in PySCIPOpt.

\hspace*{2em}Step 1: Using the correct mathematical formulation $\mathcal{MF}^*$ as a reference, first compute the size of $\mathcal{MF}^*$, $S(\mathcal{MF}^*)$, by summing the sizes of all core expressions (variables, objectives, and constraints) in $\mathcal{MF}^*$.

\hspace*{2em}Step 2: Identify which components of $\mathcal{MF}^*$ are incorrectly implemented in the PySCIPOpt code. When computing the size of the corresponding mathematical formulation, $S(\mathcal{MF}_{err})$, focus only on the correctness of each component’s logic, ignoring other errors that do not affect the logical structure. Sum the sizes of these logically incorrect or missing components to obtain $S(\mathcal{MF}_{err})$. 

\hspace*{2em}Step 3: Calculate the error ratio $\mathcal{E}$ as
\[
\mathcal{E} = \frac{S(\mathcal{MF}_{err})}{S(\mathcal{MF}^*)}.
\]

Provide detailed, step-by-step reasoning for how $S(\mathcal{MF}^*)$ is computed from the correct formulation, how $S(\mathcal{MF}_{err})$ is determined based on missing or incorrect components, and report the final numeric value of the error ratio.

\end{tcolorbox}

    \subsection{Prompt Templates for Data Synthesis}
    \label{appendix:data_synthesis_prompt}

        We use Deepseek-R1 for the error-driven reverse data synthesis pipeline.

\begin{tcolorbox}[
    enhanced, 
    colback=blue!5!white, 
    colframe=blue!50!white,
    coltitle=white,
    fonttitle=\bfseries,
    title=Prompt template used for single-error data synthesis,
    arc=6pt, 
    boxrule=1pt,
    width=\textwidth,
    sharp corners=south,
    drop shadow
]
You are a data synthesis expert in operations research. You will be given:

- A natural language description of an optimization problem.

- A correct mathematical formulation of the optimization problem.

- PySCIPOpt code that may contain errors for the optimization problem.

```$\{\texttt{question}\}$''' is the natural language description of an optimization problem.

```$\{\texttt{mathematical formulation}\}$'''  is the correct mathematical formulation of the optimization problem

```$\{\texttt{python}\}$''' is the PySCIPOpt code that may contain errors for the optimization problem.

Your task:

1. Carefully compare the PySCIPOpt code against both the natural language description and the correct mathematical formulation to detect all errors. These errors could include missing constraints, incorrect coefficients in the objective function or constraints, improper variable bounds or types (e.g., continuous instead of integer), a wrong objective direction (e.g., maximization instead of minimization), or other logical errors in translating the mathematical formulation into PySCIPOpt code.

2. Identify the specific portions of the PySCIPOpt code that are erroneous and label them as $\texttt{Error\_Code\_Portion}$. Also, identify and label the parts of the PySCIPOpt code that correctly implement the problem’s requirements as $\texttt{Correct\_Code\_Portion}$. Then, for each $\texttt{Error\_Code\_Portion}$, provide the corrected PySCIPOpt code and label it as the $\texttt{Corrected\_Code\_Portion}$. From this corrected code, explicitly define the underlying modeling logic or pattern that was initially misapplied; this will be referred to as the $\texttt{Corrected\_Modeling\_Pattern}$.

3. Based on the $\texttt{Corrected\_Modeling\_Pattern}$, generate as many distinct additional problem instances as reasonably possible. These instances should showcase variety, covering different types of optimization problems, such as assignment and resource allocation optimization, cutting and packing optimization, domain-specific optimization (e.g., specific to a particular industry), facility location optimization, financial and revenue optimization, network flow optimization, production planning and scheduling optimization, or transportation and routing optimization. Similarly, explore diverse application scenarios, including agriculture, energy, health, retail, environment, education, financial services, transportation, public utilities, manufacturing, software, construction, legal, customer service, entertainment, and others. Each generated instance must include a natural language description (in plain English), its complete mathematical formulation, and the corresponding PySCIPOpt code.

4. You must ensure that the additional problem instances generated in the previous step adhere to a critical principle of uniqueness and focused reusability. Specifically, while each new problem instance must incorporate an implementation that is analogous in its core logic to the $\texttt{Corrected\_Modeling\_Pattern}$ (this pattern can be adapted, for instance, by using a different number of variables, different coefficients suitable for the new problem within that pattern, or a moderately more complex variant of the same core idea), all other components of each new problem instance must be fundamentally different and more complex (more variables, more constraints, more advanced modeling strategies). This means the objective function, other constraints, overall problem structure, and variable sets not directly involved in the $\texttt{Corrected\_Modeling\_Pattern}$ must not resemble the $\texttt{Correct\_Code\_Portion}$ of the original PySCIPOpt code or the details of the original natural language description and correct mathematical formulation. This ensures that the additional problem instances are truly distinct from the original optimization problem in both their formulation and implementation, beyond the shared corrected modeling pattern.

5. Present the output as a JSON list of objects, each with fields “question” (problem description) and “code\_solution” (PySCIPOpt code).

\end{tcolorbox}

\begin{tcolorbox}[
    enhanced,
    colback=blue!5!white, 
    colframe=blue!50!white,
    coltitle=white,
    fonttitle=\bfseries,
    title=Prompt template used for multi-error data synthesis,
    arc=6pt, 
    boxrule=1pt,
    width=\textwidth,
    sharp corners=south,
    drop shadow
]

You are a data synthesis expert in operations research. You will be given two automated optimization modeling problems (Problem A and Problem B), each composed of three components:

- A natural language description of an optimization problem,

- A correct mathematical formulation of the  optimization problem,

- PySCIPOpt code that may contain errors for the optimization problem.

You need to identify the errors in the two automated optimization modeling problems and perform data synthesis to construct more challenging instances compared to the original problems. You can follow the steps below to do this:

1. Carefully compare the PySCIPOpt code for Problem A and Problem B against their corresponding natural language descriptions and mathematical formulations. Identify and document all discrepancies, including but not limited to: missing constraints, incorrect coefficients in the objective function or constraints, improper variable bounds or types (e.g., continuous instead of integer), a wrong objective direction (e.g., maximization instead of minimization), or other logical errors in translating the mathematical model into PySCIPOpt code.

2. Identify the specific portions of the PySCIPOpt code that are erroneous and label them as $\texttt{Error\_Code\_Portion}$ for Problem A and Problem B. Also, identify and label the parts of the PySCIPOpt code that correctly implement the problem’s requirements as $\texttt{Correct\_Code\_Portion}$ for Problem A and Problem B. Then, for each $\texttt{Error\_Code\_Portion}$, provide the corrected PySCIPOpt code, labeling it as the $\texttt{Corrected\_Code\_Portion}$ for Problem A and Problem B. From this corrected code, explicitly define the underlying modeling logic or pattern that was initially misapplied; this will be referred to as the $\texttt{Corrected\_Modeling\_Pattern}$ for Problem A and Problem B.

3. Based on the $\texttt{Corrected\_Modeling\_Pattern}$ for Problem A and Problem B, you should generate new, more complex instances that simultaneously include the $\texttt{Corrected\_Modeling\_Pattern}$ of both Problem A and Problem B within a single instance. These instances should showcase variety, covering different optimization problem types such as assignment and resource allocation optimization, cutting and packing optimization, domain-specific optimization (e.g., specific to a particular industry), facility location optimization, financial and revenue optimization, network flow optimization, production planning and scheduling optimization, or transportation and routing optimization. Similarly, explore diverse application scenarios, including agriculture, energy, health, retail, environment, education, financial services, transportation, public utilities, manufacturing, software, construction, legal, customer service, entertainment, and others. Each generated instance must include a natural language description (in plain English), its complete mathematical formulation, and the corresponding PySCIPOpt code.

4. For each newly generated instance, you must simultaneously include the $\texttt{Corrected\_Modeling\_Pattern}$ of both Problem A and Problem B. The rest of the mathematical formulation can be arbitrary, but it should be substantially different from the original formulations of Problem A and Problem B.

5. Present the output as a JSON list of objects, each with fields ``question'' (problem description) and ``code\_solution'' (PySCIPOpt code).

Automated optimization problem A as follows:


{\texttt{\{question1\}}} is the natural language description of an optimization problem. {\texttt{\{model1\}}} is the correct mathematical formulation for the optimization problem. {\texttt{\{python1\}}} is PySCIPOpt code for the optimization problem.

Automated optimization problem B as follows:


{\texttt{\{question2\}}} is the natural language description of an optimization problem. {\texttt{\{model2\}}} is a correct mathematical formulation for the optimization problem. {\texttt{\{python2\}}} is PySCIPOpt code for the optimization problem.

Now, follow the examples to present the output as a JSON list of object...

\end{tcolorbox}

    \subsection{Prompt Template for Chain-of-Thought}
    \label{appendix:chain_of_thinking_prompt}

        Following DeepSeek-R1-Zero~\citep{guo2025deepseek} and SIRL~\citep{chen2025solver}, we adopt a chain-of-thought prompt. First, we prompt the LLM to analyze the problem and extract key information to build a rationale. Second, we prompt the LLM to construct a mathematical formulation. Finally, we prompt the LLM to translate the mathematical formulation into executable PySCIPOpt Python code.

        \begin{tcolorbox}[
    enhanced, 
    colback=blue!5!white, 
    colframe=blue!50!white,
    coltitle=white,
    fonttitle=\bfseries,
    title=Prompt template used for chain-of-thought reasoning with Qwen2.5-7B-Instruct,
    arc=6pt, 
    boxrule=1pt,
    width=\textwidth,
    sharp corners=south,
    drop shadow
]
\textbf{SYSTEM:} You are a helpful assistant with expertise in mathematical modeling and the PySCIPOpt solver. When the User provides an operations research problem, you will analyze it, build a detailed mathematical model, and provide the PySCIPOpt code to solve it.

Your response should follow these steps:
\begin{enumerate}
    \item \texttt{<think>}  
    
    Carefully analyze the problem to identify decision variables, objective, and constraints. 
    
    \texttt{</think>}
    \item \texttt{<model>} 
    
    Develop a complete mathematical model, explicitly defining:
    \begin{itemize}
        \item Sets
        \item Parameters
        \item Decision Variables (and their types)
        \item Objective Function
        \item Constraints
    \end{itemize}
    
    \texttt{</model>}
    \item \texttt{<python>} 
    
    Provide the corresponding PySCIPOpt Python code to implement the model. 
    
    \texttt{</python>}
\end{enumerate}

\textbf{USER:} Answer the following mathematical modeling question:

```question

\texttt{\{question\}}

'''

Let\'s think step by step and fill in the PySCIPOpt code into

```
python

\texttt{\{python\}}

'''.

\end{tcolorbox}

        \begin{tcolorbox}[
    enhanced, 
    colback=blue!5!white, 
    colframe=blue!50!white,
    coltitle=white,
    fonttitle=\bfseries,
    title=Prompt template used for chain-of-thought reasoning with Qwen3-8B,
    arc=6pt, 
    boxrule=1pt,
    width=\textwidth,
    sharp corners=south,
    drop shadow
]
\textbf{SYSTEM:} You are a helpful assistant with expertise in mathematical modeling and the PySCIPOpt solver. When the User provides an operations research problem, you will analyze it, build a detailed mathematical model, and provide the PySCIPOpt code to solve it.

Your response should follow these steps:
\begin{enumerate}
    \item \texttt{<analysis>}  
    
    Carefully analyze the problem to identify decision variables, objective, and constraints. 
    
    \texttt{</analysis>}
    \item \texttt{<model>} 
    
    Develop a complete mathematical model, explicitly defining:
    \begin{itemize}
        \item Sets
        \item Parameters
        \item Decision Variables (and their types)
        \item Objective Function
        \item Constraints
    \end{itemize}
    
    \texttt{</model>}
    \item \texttt{<python>} 
    
    Provide the corresponding PySCIPOpt Python code to implement the model. 
    
    \texttt{</python>} 
\end{enumerate}

\textbf{USER:} Answer the following mathematical modeling question:

```question

\texttt{\{question\}}

'''

Let\'s think step by step and fill in the PySCIPOpt code into

```
python

\texttt{\{python\}}

'''. /no\_think

\end{tcolorbox}

    \subsection{Prompt Template for Dynamic SFT}
    \label{appendix:dynamic_sft}

        First, we design a prompt to generate \texttt{correct\_response}. Specifically, we use ground-truth solutions as guidance and independently solve the operations research problems through chain-of-thought reasoning, thereby generating the desired responses.

        \begin{tcolorbox}[
    enhanced, 
    colback=blue!5!white, 
    colframe=blue!50!white,
    coltitle=white,
    fonttitle=\bfseries,
    title=Prompt template used to generate correct response,
    arc=6pt, 
    boxrule=1pt,
    width=\textwidth,
    sharp corners=south,
    drop shadow
]

\textbf{SYSTEM:} You are a helpful Assistant with expertise in mathematical modeling and the PySCIPOpt solver. 
When the User provides an OR question, you will analyze it, build a detailed mathematical model, 
and provide the PySCIPOpt code to solve it.

Before answering, you may review the provided reference reasoning or code $\texttt{\{ground\_truth\_formulation\}}$ for guidance only. 
Do not copy or rely on it directly. Your solution must be fully generated independently, using your own analysis and reasoning.

    Your response should follow these steps:
    
    1. \texttt{<analysis>}
    
       \hspace*{2em}Explain how the reference $\texttt{\{ground\_truth\_formulation\}}$ can guide your reasoning.
       Highlight any insights or techniques you can borrow, 
       but do not copy any content verbatim.
       Be concise and structured.
       
       \hspace*{1em}\texttt{</analysis>}
       
    2. \texttt{<response>}
    
        \hspace*{2em}Provide your complete independent solution, including:
        \begin{enumerate}
            \item \texttt{<think>}  
            
            Carefully analyze the problem to identify decision variables, objective, and constraints. 
            
            \texttt{</think>}
            \item \texttt{<model>} 
            
            Develop a complete mathematical model, explicitly defining:
            \begin{itemize}
                \item Sets
                \item Parameters
                \item Decision Variables (and their types)
                \item Objective Function
                \item Constraints
            \end{itemize}
            
            \texttt{</model>}
            \item \texttt{<python>} 
            
            Provide the corresponding PySCIPOpt Python code to implement the model. 
            
            \texttt{</python>}
        \end{enumerate}
        
       \hspace*{1em}\texttt{</response>}

Your final output must therefore contain exactly two sections:

\texttt{<analysis>}...\texttt{</analysis>}

\texttt{<response>}...\texttt{</response>}

\textbf{USER:} Answer the following mathematical modeling question:

```question

\texttt{\{question\}}

'''

Let\'s think step by step.

\end{tcolorbox}

        Then, we design a prompt to correct wrong responses. Specifically, we use $\texttt{correct\_response}$ as a reference to correct wrong responses from LLM post-training rollouts, thereby obtaining the corrected responses.

        \begin{tcolorbox}[
    enhanced, 
    colback=blue!5!white, 
    colframe=blue!50!white,
    coltitle=white,
    fonttitle=\bfseries,
    title=Prompt template used to correct wrong response,
    arc=6pt,
    boxrule=1pt,
    width=\textwidth,
    sharp corners=south,
    drop shadow
]

You are a helpful assistant with expertise in mathematical modeling and the PySCIPOpt solver.
The operations research question is as follows:

$\texttt{\{question\}}$.

The correct mathematical modeling response (for reference only) is as follows:

$\texttt{\{correct\_response\}}$.

The wrong mathematical modeling response from another LLM is as follows:

$\texttt{\{wrong\_response\}}$.

Your task:

1. Write your reasoning about how to modify the wrong response based on the correct response inside \texttt{<analysis>}...\texttt{</analysis>} tags.

    \hspace*{1em}- In this section you may explain which parts of the wrong response are incorrect, why, and how they should be corrected.
    
    \hspace*{1em}- Be concise and structured.
    
2. Output the **entire corrected version of the wrong response** inside \texttt{<corrected response>}...\texttt{</corrected response>} tags.

   \hspace*{1em}- The corrected response must preserve all parts of the wrong response that are already correct.
   
   \hspace*{1em}- Change only the portions that are actually incorrect.
   
   \hspace*{1em}- Do not add extra explanation, justification, or commentary in this section — only the corrected content.
   
   \hspace*{1em}- Keep the same Python coding style as in the wrong response. Do not wrap code into a function.

Your final output must therefore contain exactly two sections:

\texttt{<analysis>}...\texttt{</analysis>}

\texttt{<corrected response>}...\texttt{</corrected response>}

\end{tcolorbox}

\section{MIND Details}
\label{appendix:mind_details}
    
    \subsection{Training and Inference Details}
\label{appendix:training_details}

    \paragraph{Training Hyperparameters}
    
        All experiments were conducted on a single computing node equipped with four NVIDIA A100 GPUs, each with 80 GB of memory. The ms-swift framework~\citep{zhao2025swift} was used to implement SFT, while the VeRL framework~\citep{sheng2025hybridflow} was used to implement GRPO, DAPO and DFPO. All training hyperparameters are listed in Table~\ref{tab:dfpo_training_configuration}, Table~\ref{tab:dapo_training_configuration}, Table~\ref{tab:grpo_training_configuration} and Table~\ref{tab:sft_training_configuration}.

        \begin{table*}[htbp!]
    \centering
    \caption{List of training hyperparameters and their values used in the DFPO.}
    \label{tab:dfpo_training_configuration}
    \begin{tabular}{ll}
    \toprule
    \toprule
    \multicolumn{2}{c}{\textbf{Data}} \\
    \midrule
    \textbf{Parameter} & \textbf{Value} \\
    \midrule
    Optimizer & AdamW \\
    Training epochs & 26 \\
    Training batch size & 1024 \\
    Max prompt length & 4096 \\
    Max response length  & 8192 \\
    Learning rates & $10^{-6}$  \\
    Truncation & left  \\
    \midrule
    \midrule
    \multicolumn{2}{c}{\textbf{Actor}} \\
    \midrule
    \midrule
    \textbf{Parameter} & \textbf{Value} \\
    \midrule
    Number of rollouts per prompt & 8 \\
    PPO mini-batch size & 256  \\
    Clip ratio low  & 0.20 \\  
    Clip ratio high & 0.28  \\
    Entropy loss & Disabled  \\
    KL loss & Disabled  \\
    Gradient clipping & 1.0  \\
    temperature (sampling) & 1.0  \\
    Top p (sampling) & 1.0  \\
    Top k (sampling) & -1  \\
    $\alpha$ & 0.2  \\
    $\beta$ & 0.05  \\
    $\gamma$ & 0.8  \\
    \midrule
    \midrule
    \multicolumn{2}{c}{\textbf{Reward}} \\
    \midrule
    \midrule
    \textbf{Parameter} & \textbf{Value} \\
    \midrule
    Overlong buffer length & 4096  \\
    Overlong penalty factor  & 1.0 \\  
    \bottomrule
    \bottomrule
    \end{tabular}
\end{table*}

\begin{table*}[htbp!]
    \centering
    \caption{List of training hyperparameters and their values used in the DAPO.}
    \label{tab:dapo_training_configuration}
    \begin{tabular}{ll}
    \toprule
    \toprule
    \multicolumn{2}{c}{\textbf{Data}} \\
    \midrule
    \textbf{Parameter} & \textbf{Value} \\
    \midrule
    Optimizer & AdamW \\
    Training epochs & 26 \\
    Training batch size & 1024 \\
    Max prompt length & 4096 \\
    Max response length  & 8192 \\
    Learning rates & $10^{-6}$  \\
    Truncation & left  \\
    \midrule
    \midrule
    \multicolumn{2}{c}{\textbf{Actor}} \\
    \midrule
    \midrule
    \textbf{Parameter} & \textbf{Value} \\
    \midrule
    Number of rollouts per prompt & 8 \\
    PPO mini-batch size & 256  \\
    Clip ratio low  & 0.20 \\  
    Clip ratio high & 0.28  \\
    Entropy loss & Disabled  \\
    KL loss & Disabled  \\
    Gradient clipping & 1.0  \\
    temperature (sampling) & 1.0  \\
    Top p (sampling) & 1.0  \\
    Top k (sampling) & -1  \\
    \midrule
    \midrule
    \multicolumn{2}{c}{\textbf{Reward}} \\
    \midrule
    \midrule
    \textbf{Parameter} & \textbf{Value} \\
    \midrule
    Overlong buffer length & 4096  \\
    Overlong penalty factor  & 1.0 \\  
    \bottomrule
    \bottomrule
    \end{tabular}
\end{table*}

\begin{table*}[htbp!]
    \centering
    \caption{List of training hyperparameters and their values used in the GRPO.}
    \label{tab:grpo_training_configuration}
    \begin{tabular}{ll}
    \toprule
    \toprule
    \multicolumn{2}{c}{\textbf{Data}} \\
    \midrule
    \textbf{Parameter} & \textbf{Value} \\
    \midrule
    Optimizer & AdamW \\
    Training epochs & 26 \\
    Training batch size & 1024 \\
    Max prompt length & 2048 \\
    Max response length  & 8192 \\
    Learning rates & $10^{-6}$  \\
    \midrule
    \midrule
    \multicolumn{2}{c}{\textbf{Actor}} \\
    \midrule
    \midrule
    \textbf{Parameter} & \textbf{Value} \\
    \midrule
    Number of rollouts per prompt & 8 \\
    PPO mini-batch size & 256  \\
    Entropy loss & Disabled  \\
    KL loss coefficient & 0.001  \\
    KL loss type & Low Var KL  \\
    Gradient clipping & 1.0  \\
    temperature (sampling) & 1.0  \\
    Top p (sampling) & 1.0  \\
    Top k (sampling) & -1  \\
    \bottomrule
    \bottomrule
    \end{tabular}
\end{table*}

\begin{table*}[htbp!]
    \centering
    \caption{List of training hyperparameters and their values used in the SFT.}
    \label{tab:sft_training_configuration}
    \begin{tabular}{ll}
    \toprule
    \toprule
    \textbf{Parameter} & \textbf{Value} \\
    \midrule
    Optimizer & AdamW \\
    Training epochs & 3 \\
    Training batch size & 2 \\
    Gradient accumulation steps & 8 \\
    Max prompt length & 4096 \\
    Max response length  & 8192 \\
    Learning rates & $10^{-4}$  \\
    Train type & LoRA~\cite{yu2023low}  \\
    LoRA rank  & 8 \\  
    LoRA alpha & 32 \\ 
    \bottomrule
    \bottomrule
    \end{tabular}
\end{table*}

    \paragraph{Inference Hyperparameters}

        As shown in Table~\ref{tab:inference_configuration}, we use a greedy decoding strategy for LLM inference to ensure reproducibility.

\begin{table}[htbp!]
    \centering
    \caption{List of inference hyperparameters and their values used in the DFPO.}
    \label{tab:inference_configuration}
    \begin{tabular}{ll}
    \toprule
    \toprule
    \multicolumn{2}{c}{\textbf{Decoding Settings}} \\
    \midrule
    \textbf{Parameter} & \textbf{Value} \\
    \midrule
    Max tokens & 8192 \\
    Temperature & 0.0 \\
    \bottomrule
    \bottomrule
    \end{tabular}
\end{table}

\subsection{Preliminary Results on Deepseek-V3}

\begin{figure}
        \centering
        \includegraphics[width=0.5\linewidth]{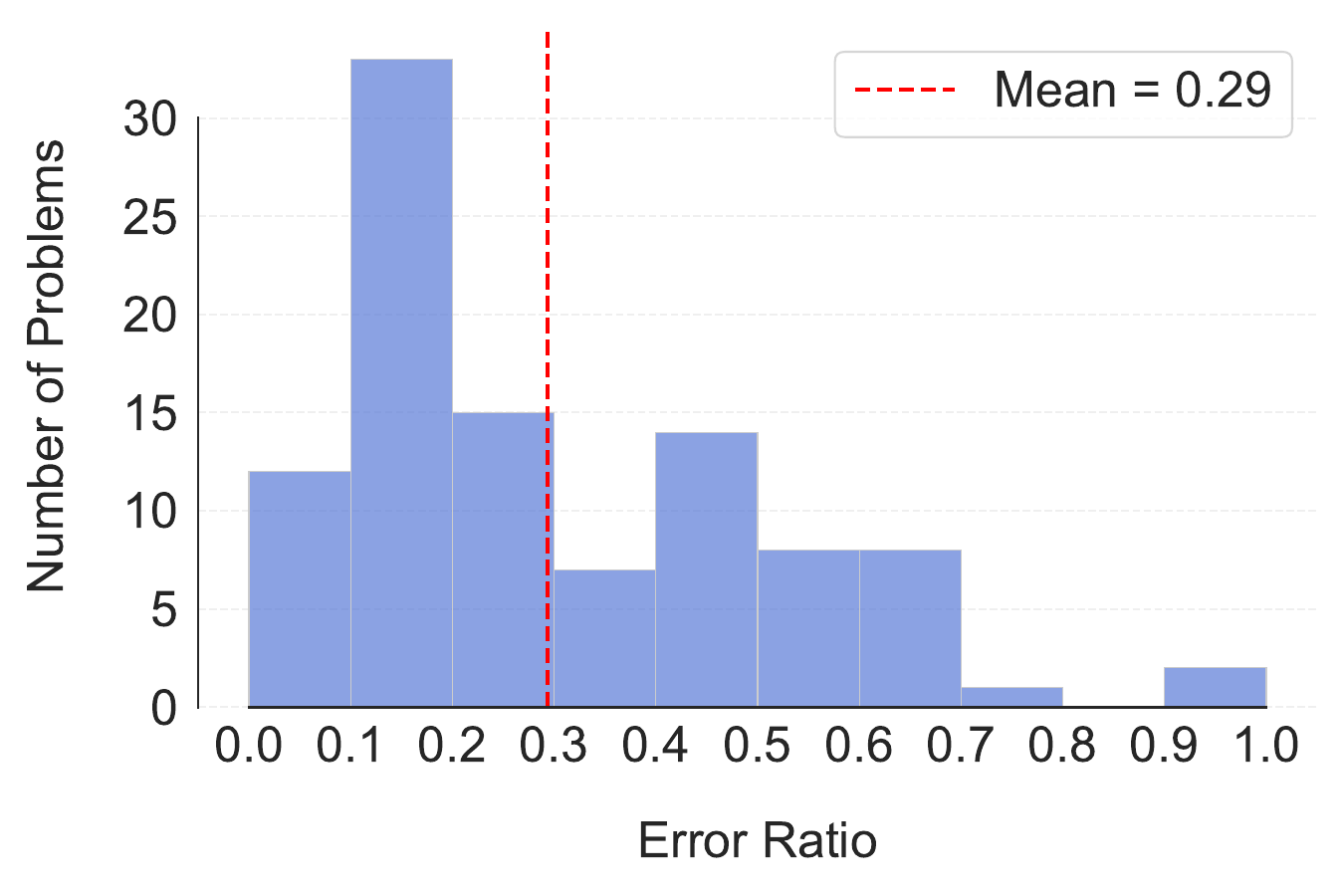}
        \caption{Distribution of error ratio across 100 incorrect generation results for Deepseek-V3.}
        \label{fig:deepseek_error_ratio_distribution}
    \end{figure}

    As a supplement to the preliminary results on Qwen2.5-7B-Instruct, we conduct the same preliminary experiments using Deepseek-V3, a model with a different architecture, on the OR-Instruct-3K. We also analyze the distribution of error ratios for the questions on which Deepseek-V3 make errors. As shown in Figure~\ref{fig:deepseek_error_ratio_distribution}, when errors occur, Deepseek-V3 also introduces only a small fraction of errors rather than producing entirely incorrect formulations in most cases, further supporting the conclusions observed for Qwen2.5-7B-Instruct. Additionally, we find that Deepseek-V3 has a lower average error ratio of 29\% compared with 33\% for Qwen2.5-7B-Instruct, indicating that more powerful LLM may have a higher capacity to produce fewer errors per instance.

\subsection{Reward Weight Sensitivity Analysis}
\label{appendix:alpha_sensitivity_analysis}

\begin{figure}
        \centering
        \includegraphics[width=1.0\linewidth]{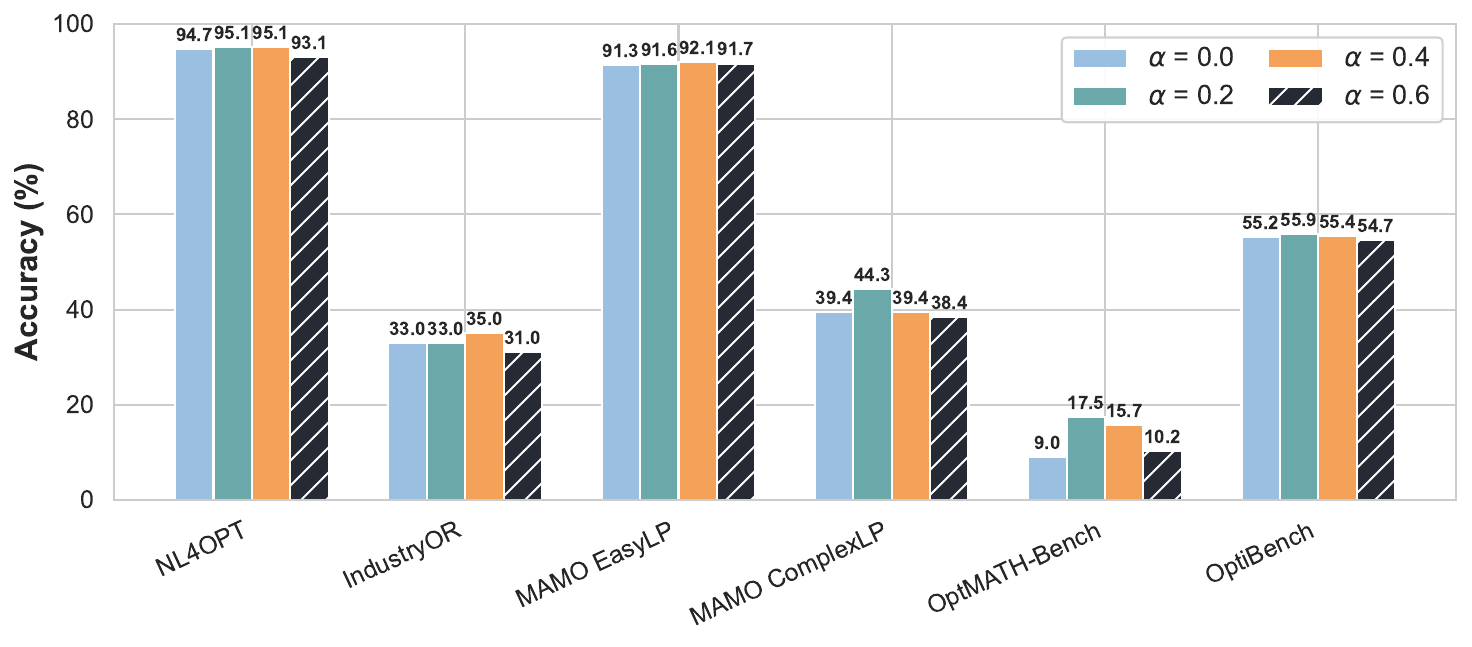}
        \caption{Performance comparison for different $\alpha$ in the reward function.}
        \label{fig:alpha_sensitivity_analysis}
    \end{figure}

    For our reward function hyperparameter $\alpha$, we evaluate its influence by testing values in \{0.0, 0.2, 0.4, 0.6\}, with the results shown in Figure~\ref{fig:alpha_sensitivity_analysis}. For the experimental details, we use DAPO to train Qwen2.5-7B-Instruct on the training dataset (10,000 instances) for 7 epochs. We note that $\alpha=0.0$ corresponds to a standard 0-1 reward. The results show that $\alpha=0.2$ and $\alpha=0.4$ achieve better performance on most benchmarks compared with $\alpha=0.0$ and $\alpha=0.6$, indicating that the fidelity reward, as an auxiliary signal, should not dominate the final reward value.

\subsection{Ablation Study of Data Synthesis Strategies}
\label{appendix:decomposition_ablation_study}

\begin{figure}
        \centering
        \includegraphics[width=0.7\linewidth]{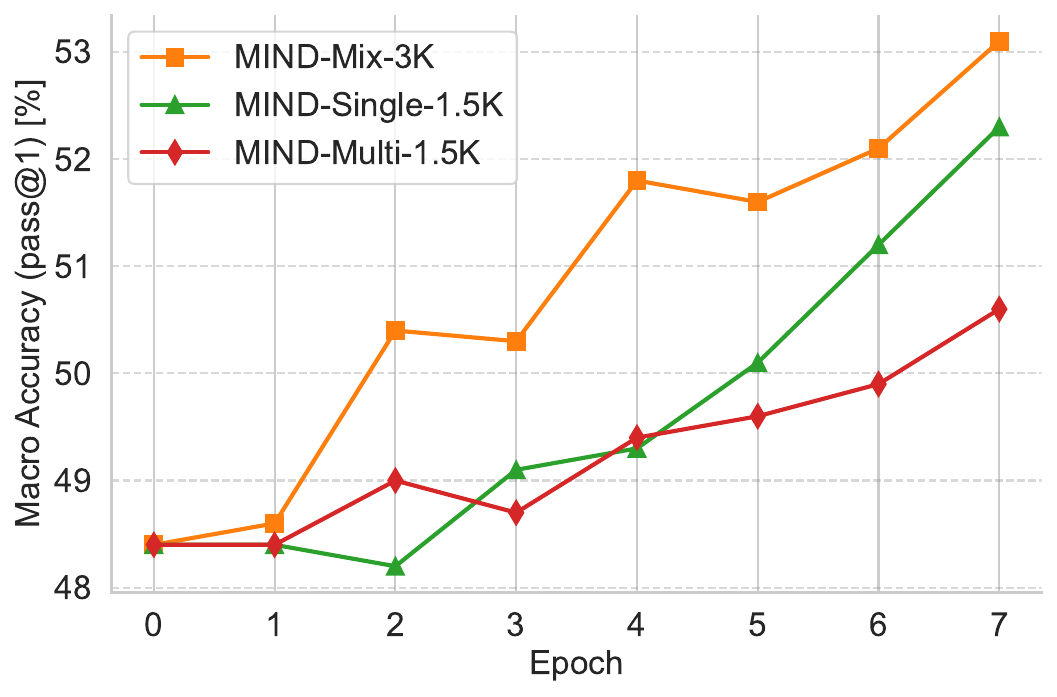}
        \caption{Ablation study for the single-error and multi-error strategies across six benchmarks.}
        \label{fig:decomposition_ablation_study}
    \end{figure}

    To verify the difference between single-error and multi-error strategies, we split the MIND-3K training dataset, which is a mixture of single-error and multi-error data synthesis dataset, into MIND-Single-1.5K (1,500 instances) and MIND-Multi-1.5K (1,500 instances). We then employ DAPO to train Qwen2.5-7B-Instruct from scratch on MIND-Single-1.5K and MIND-Multi-1.5K. As shown in Figure~\ref{fig:decomposition_ablation_study}, the model achieves better training performance on MIND-Mix-3K compared with MIND-Single-1.5K and MIND-Multi-1.5K. Furthermore, Table~\ref{tab:error_stategies_ablation_study_table} presents a detailed performance comparison after seven training epochs across six benchmarks. Our results also show that training on MIND-Single-1.5K leads to better performance than training on MIND-Multi-1.5K. We hypothesize that this disparity arises because LLMs struggle to learn effectively when trained directly on highly challenging datasets. To further substantiate this hypothesis, we evaluate Qwen2.5-7B-Instruct on both datasets. The model achieves an average accuracy of 52.9\% on MIND-Single-1.5K, but only 41.2\% on MIND-Multi-1.5K. This pronounced accuracy gap corroborates our claim that multi-error reverse data synthesis generates datasets that are substantially more difficult than those produced by single-error synthesis.

    \begin{table}[htbp!]
    \centering
    \caption{Ablation results for the single-error and multi-error strategies on Qwen2.5-7B-Instruct. (pass@1$\uparrow$).}
    \label{tab:error_stategies_ablation_study_table}
    \resizebox{1.0\textwidth}{!}{
        \begin{tabular}{cccccccc}
        \toprule
        \textbf{Data} & \textbf{NL4OPT} & \textbf{IndustryOR} & \textbf{EasyLP} & \textbf{ComplexLP}  & \textbf{OptMATH} & \textbf{OptiBench} & \textbf{Macro AVG} \\
        \midrule
        MIND-Single-1.5K & 91.4\% & 29.0\% & 90.4\% & 40.9\% & 8.4\% & 53.6\% & 52.3\% \\
        MIND-Multi-1.5K & 91.4\% & 29.0\% & 90.2\% & 33.0\% & 6.0\% & 54.0\% & 50.6\% \\
        MIND-Mix-3K & 94.3\% & 30.0\% & 90.8\% & 39.9\% & 7.8\% & 55.5\% & 53.1\% \\
        \bottomrule
        \end{tabular}
    }
\end{table}

\subsection{Modeling Error Analysis}
\label{appendix:modeling_error_analysis}

    \begin{table}[t]
    \centering
    \caption{Analysis of modeling error types in optimization modeling.}
    \renewcommand{\arraystretch}{1.3}
    \begin{tabularx}{\linewidth}{l X c c}
    \toprule
    \toprule
    \textbf{Error Type} & \textbf{Concrete Error} & \textbf{Qwen2.5-7B} & \textbf{MIND-Qwen2.5-7B} \\
    \midrule
    \midrule
    \multirow{4}{*}{Variables} 
    & Incorrect decision variables. & 12.1\%   & 15.5\% \textcolor{green}{($\uparrow$ 3.4\%)} \\
    \cline{2-4} 
    & Decision variables omission. & 4.4\%   &  10.6\% \textcolor{green}{($\uparrow$ 6.2\%)} \\
    \cline{2-4} 
    & Superfluous decision variables. & 7.7\%   &  8.1\% \textcolor{green}{($\uparrow$ 0.4\%)} \\
    \cline{2-4} 
    & Incorrect variable types. & 11.8\%   & 7.1\% \textcolor{red}{($\downarrow$ 4.7\%)} \\
    \midrule
    \multirow{9}{*}{Objective} 
    & Optimization direction error. & 1.4\%   & 0.0\% \textcolor{red}{($\downarrow$ 1.4\%)} \\
    \cline{2-4} 
    & Incorrect objective terms. & 12.8\%   & 4.2\% \textcolor{red}{($\downarrow$ 8.6\%)} \\
    \cline{2-4} 
    & Objective terms omission. & 3.0\%   & 2.5\% \textcolor{red}{($\downarrow$ 0.5\%)} \\
    \cline{2-4} 
    & Superfluous objective terms. & 1.7\%   & 0.4\% \textcolor{red}{($\downarrow$ 1.3\%)} \\
    \cline{2-4}
    & Incorrect or missing advanced modeling techniques. The incorrect application or omission of sophisticated modeling techniques, which can lead to improper handling of multi-objective problems, non-linear objectives or other advanced modeling scenarios. & 2.7\%   & 5.3\% \textcolor{green}{($\uparrow$ 2.6\%)} \\
    \midrule
    \multirow{9}{*}{Constraints}
    & Incorrect constraint. & 11.8\%   & 15.5\% \textcolor{green}{($\uparrow$ 3.7\%)} \\
    \cline{2-4}
    & Constraint omission.  & 10.1\%   & 8.5\% \textcolor{red}{($\downarrow$ 1.6\%)} \\ 
    \cline{2-4}
    & Superfluous constraints. & 3.7\%   & 0.0\% \textcolor{red}{($\downarrow$ 3.7\%)} \\
    \cline{2-4}
    & Equality and inequality constraints confusion. & 4.0\%   & 4.2\% \textcolor{green}{($\uparrow$ 0.2\%)} \\
    \cline{2-4}
    & Incorrect or missing advanced modeling techniques. The incorrect application or omission of sophisticated modeling techniques, which can lead to improper handling of non-linear constraints, logical constraints, or other advanced modeling scenarios. & 1.0\%   & 11.7\% \textcolor{green}{($\uparrow$ 10.7\%)} \\
    \midrule
    \multirow{6}{*}{Parameters} 
    & Incorrect parameters definition. This includes missing essential parameters, incorrectly defined parameters, parameters assigned with wrong numerical values, or other incorrect parameter definition scenarios. & 8.4\%   & 4.6\% \textcolor{red}{($\downarrow$ 3.8\%)} \\
    \cline{2-4}
    & Parameters misuse. The incorrect use of defined parameters, such as value misuse, unit or scale misuse, reference errors, or other improper applications of parameters.  & 3.4\%   & 1.8\% \textcolor{red}{($\downarrow$ 1.6\%)} \\
    \bottomrule
    \bottomrule
    \end{tabularx}
    \label{tab:error_analysis}
\end{table}

    We randomly sample 300 erroneous responses each from Qwen2.5-7B-Instruct (before post-training) and MIND-Qwen2.5-7B (after DFPO-based post-training). We first defined a taxonomy of error types relevant to optimization modeling. For each query-response pair, three domain experts independently annotated the dominant error category, achieving high inter-annotator agreement. As shown in Table~\ref{tab:error_analysis}, the top five error types for Qwen2.5-7B-Instruct are ``incorrect objective terms (12.8\%)'', ``incorrect decision variables (12.1\%)'', ``incorrect constraint (11.8\%)'', ``incorrect variable types (11.8\%)'', and ``constraint omission (10.1\%)''. In contrast, the top five errors for MIND-Qwen2.5-7B are ``incorrect decision variables (15.5\%)'', ``incorrect constraint (15.5\%)'', ``incorrect or missing advanced modeling techniques (11.7\%)'', ``decision variable omission (10.6\%)'', and ``constraint omission (8.5\%)''. 
    
    Notably, while basic syntactic or structural errors (e.g., wrong variable types) diminish after post-training, new dominant errors involve more sophisticated modeling challenges, such as the appropriate use of advanced techniques (e.g., piecewise linearization, or indicator constraints) and comprehensive problem scoping (e.g., omitting key variables or constraints in complex scenarios). This shift strongly suggests that DFPO effectively mitigates simpler, surface-level errors, push the model's failure modes toward deeper, semantics-rich challenges—a hallmark of improved reasoning capability.

    \begin{figure}[htbp!]
        \centering
        \includegraphics[width=1.0\linewidth]{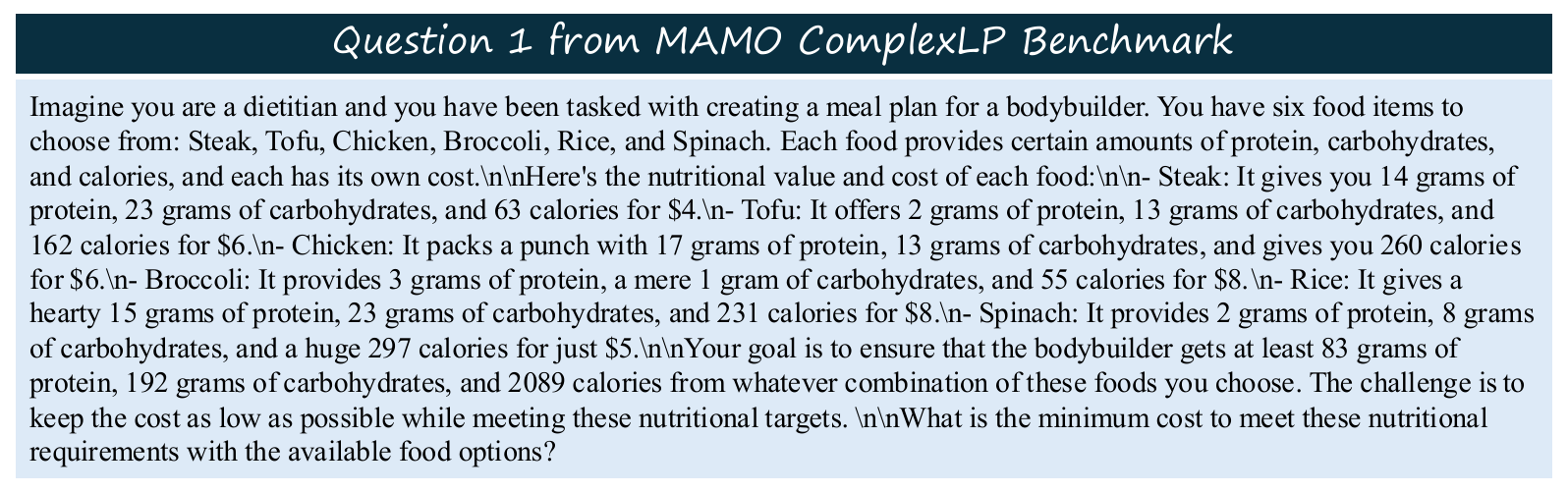}
        \caption{Question of example 1.}
        \label{fig:dynamic_question_2}
    \end{figure}

    \begin{figure}[htbp!]
        \centering
        \includegraphics[width=1.0\linewidth]{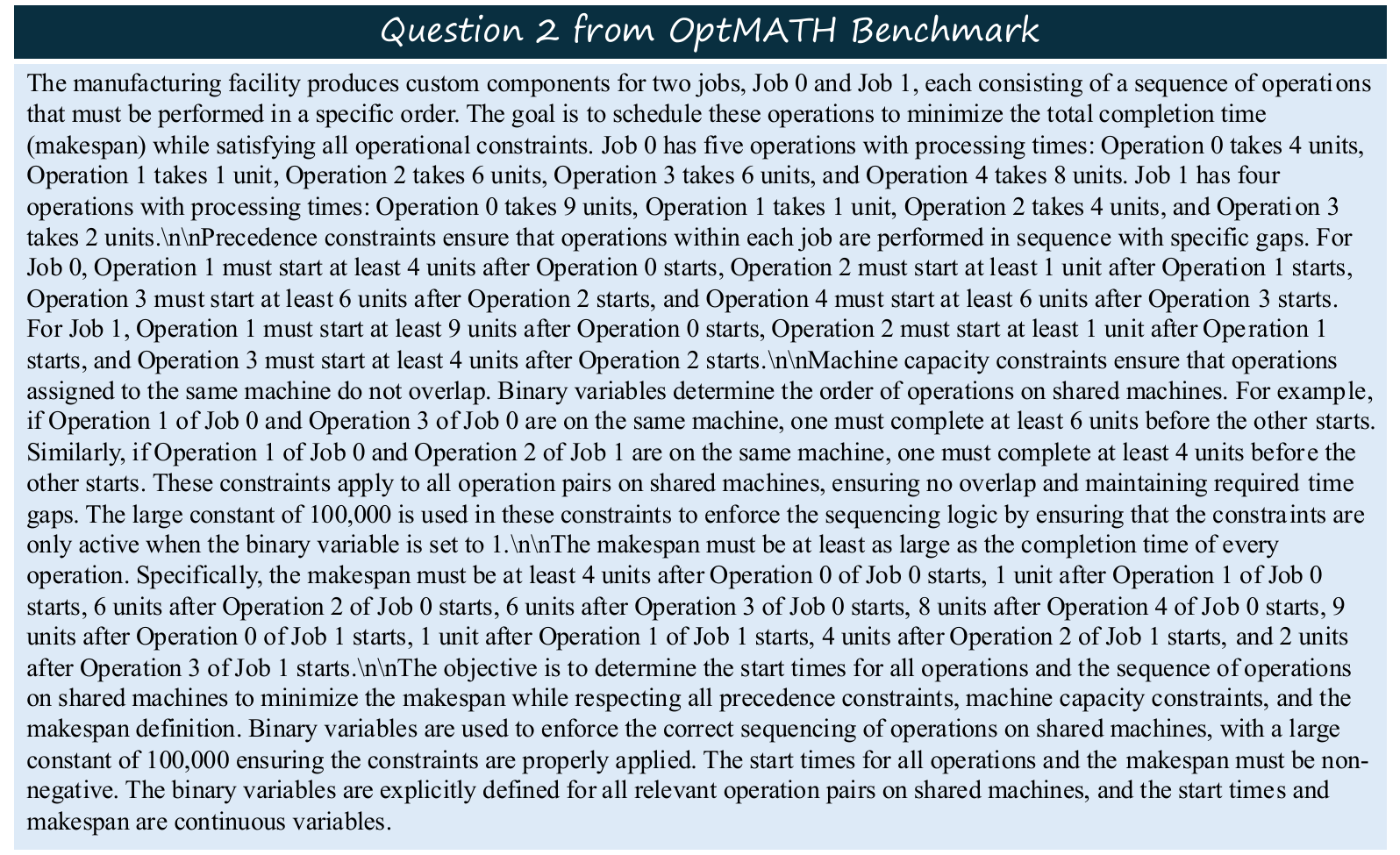}
        \caption{Question of example 2.}
        \label{fig:dynamic_question_3}
    \end{figure}

    \begin{figure}
        \centering
        \includegraphics[width=0.95\linewidth]{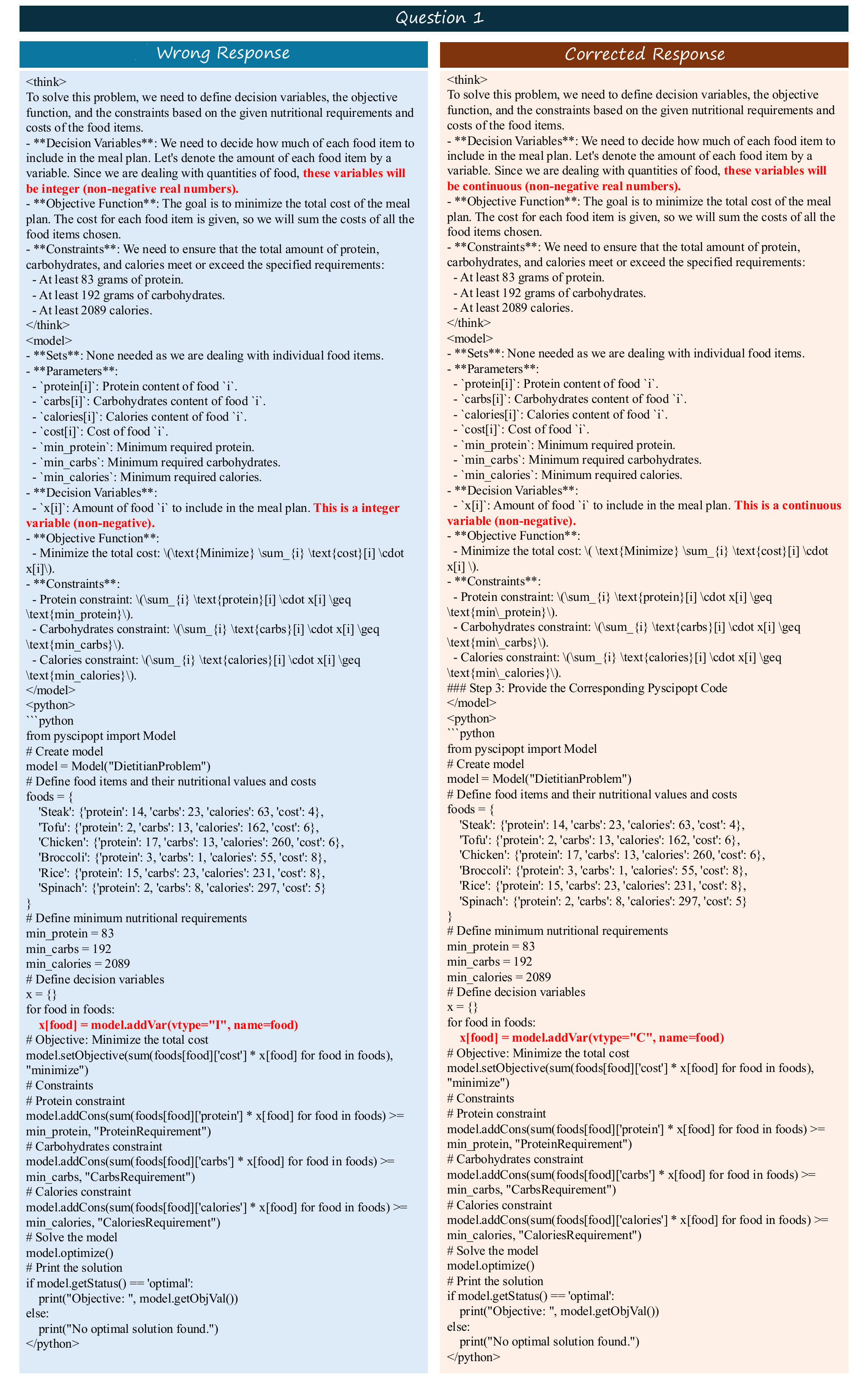}
        \caption{Wrong resposne and corrected response corresponding to example 1.}
        \label{fig:dynamic_sft_2}
    \end{figure}

    \begin{figure}
        \centering
        \includegraphics[width=0.9\linewidth]{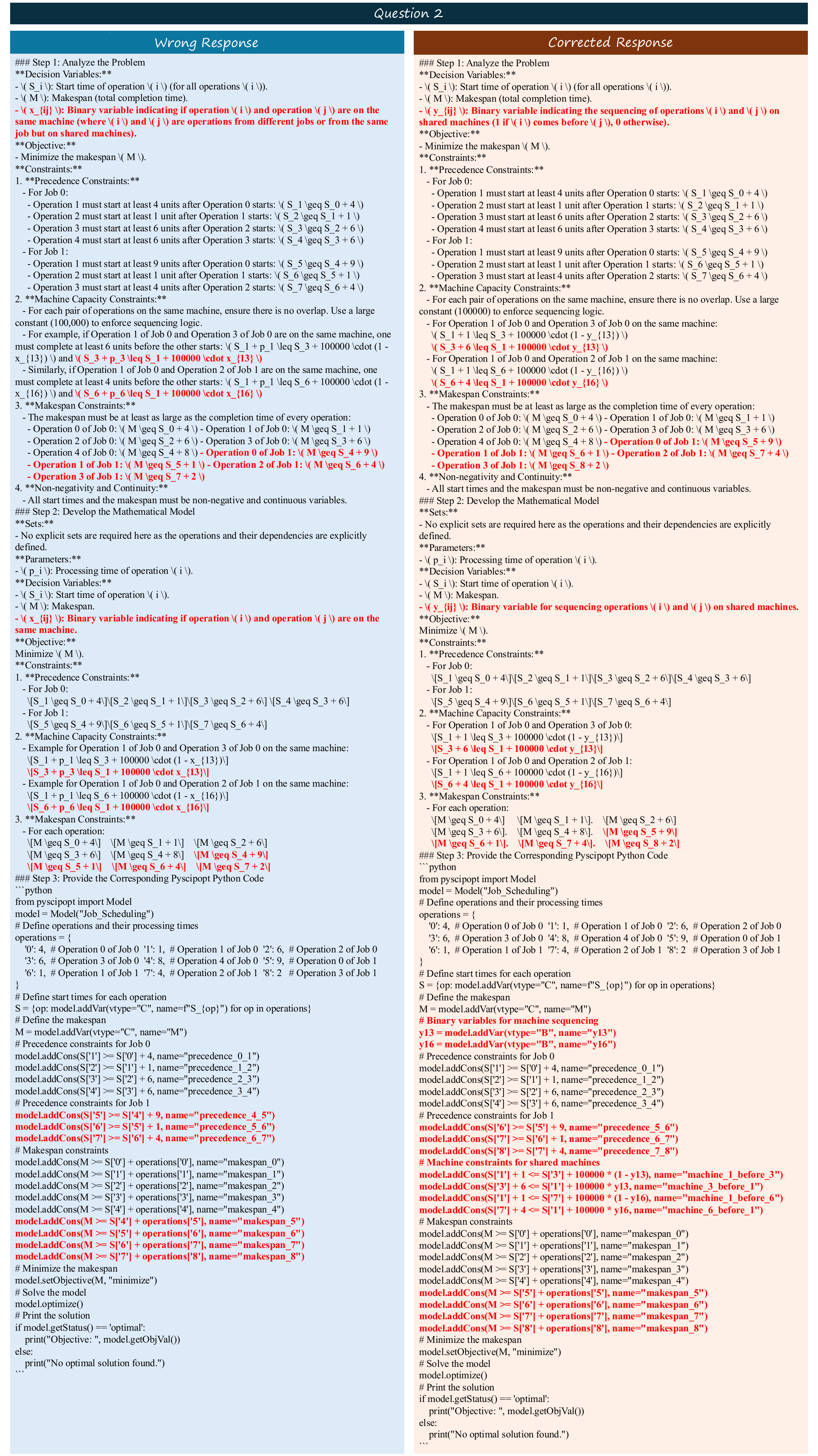}
        \caption{Wrong response and corrected response corresponding to example 2.}
        \label{fig:dynamic_sft_3}
    \end{figure}

\subsection{Case study: Examples of Corrected Wrong Responses for Dynamic SFT}
\label{appendix:dynamic_sft_examples}

    We select the first instances from MAMO ComplexLP and OptMATH as example 1 (Figure~\ref{fig:dynamic_question_2}) and example 2 (Figure~\ref{fig:dynamic_question_3}), respectively. As shown in Figure~\ref{fig:dynamic_sft_2} and Figure~\ref{fig:dynamic_sft_3}, these cases illustrate how a powerful LLM can correct the errors in the wrong response from the base model, producing a corrected response whose distribution closely matches that of the wrong response.

    Example 1 is a diet problem. Since the problem does not explicitly require food to be purchased in whole units, the variables should be allowed to take continuous values. Although the variable type is incorrect, this does not affect the correctness of the parameters, objective function, or constraints. Therefore, the errors in the wrong response are localizable. If we fix only these localizable errors, the overall answer will be corrected.

    Example 2 is a scheduling problem. The wrong response includes precedence constraints, machine capacity constraints, makespan constraints, and non-nagativity constraints. We observe minor errors in the machine capacity and makespan constraints. However, the errors in the makespan constraints do not affect the correctness of the precedence constraints, machine capacity constraints, or non-negativity constraints. Similarly, the errors in the machine capacity constraints do not affect the correctness of the precedence constraints, makespan constraints, or non-negativity constraints. Therefore, the errors in the wrong response are localizable. By fixing only these localizable errors, the overall solution can be corrected.

\end{document}